\documentclass[11pt]{article}
\usepackage[margin=1in]{geometry}
\usepackage{times}
\usepackage{algorithm}
\usepackage{algpseudocode}
\usepackage{microtype}
\usepackage{graphicx}
\usepackage{subcaption}
\usepackage{natbib}
\usepackage{booktabs}
\usepackage{makecell}
\usepackage{multirow}
\usepackage{amssymb}
\usepackage{array}
\usepackage{tikz}
\usepackage{pgfplots}
\usepgfplotslibrary{groupplots}
\pgfplotsset{compat=newest}
\usepackage{amsmath}
\usepackage{amsthm}
\usepackage{mathtools}
\usepackage{hyperref}
\usepackage[capitalize,noabbrev]{cleveref}
\usepackage{url}
\usepackage{authblk}
\usepackage{tabularx}
\usepackage{array}

\newcolumntype{C}{>{\centering\arraybackslash}X}

\newcommand{\vek}[1]{#1}

\newcommand{\dd}{\mathrm{d}}
\newcommand{\grad}[2][]{\nabla_{#1} #2 }

\newcommand{\norm}[1]{\left\lVert #1 \right\lVert}
\newcommand{\twoNorm}[1]{\norm{#1}_2}

\newcommand{\normal}[1]{ \mathcal{N} \left( {#1} \right)}

\newcommand{\param}{\theta}
\newcommand{\energy}{E_{\param}}
\newcommand{\energyi}{E_{\param_i}}
\newcommand{\score}{s_{\param}}
\newcommand{\pEbm}{p_{\param}}
\newcommand{\pEbmi}{p_{\param_i}}
\newcommand{\x}{\vek{x}}

\newcommand{\partFn}{Z_{\param}}

\newcommand{\gradX}{\grad[\x]}

\newcommand{\eps}{\vek{\epsilon}}
\newcommand{\epsP}{\eps_{\param}}
\newcommand{\epsPi}{\eps_{\param_i}}

\newcommand{\qProd}{q^{\Pi}}
\newcommand{\pProd}{\pEbm^{\Pi}}
\newcommand{\eProd}{\energy^{\Pi}}

\newcommand{\pF}{\varphi}
\newcommand{\pCfull}{p_{\pF}}

\newcommand{\langStep}{\delta}
\newcommand{\mcSteps}{L_t}
\newcommand{\mcKern}[1]{k_t \left( #1 \right)}
\newcommand{\acc}{\alpha}
\newcommand{\xCand}{\hat{\x}}

\theoremstyle{plain}

\theoremstyle{definition}

\theoremstyle{remark}

\title{MCMC-Correction of Score-Based Diffusion Models for Model Composition}

\author[*1,2]{Anders Sjöberg}
\author[*2]{Jakob Lindqvist}
\author[1]{Magnus Önnheim}
\author[1]{Mats Jirstrand}
\author[2]{Lennart Svensson}

\affil[1]{Fraunhofer-Chalmers Centre, Gothenburg, SE-412 88, Sweden}
\affil[2]{Department of Electrical Engineering, Chalmers University of Technology, Gothenburg, SE-412 96, Sweden}
\affil[*]{Equal contribution}

\date{}

\begin{document}

\maketitle
\begin{center}
\textit{Published in Entropy (2026), 28(3):351. DOI: \href{https://doi.org/10.3390/e28030351}{10.3390/e28030351}}
\end{center}

\begin{abstract}
\noindent Diffusion models can be parameterized in terms of either score or energy function. The energy parameterization is attractive as it enables sampling procedures such as Markov Chain Monte Carlo (MCMC) that incorporates a Metropolis--Hastings (MH) correction step based on energy differences between proposed samples. Such corrections can significantly improve sampling quality, particularly in the context of model composition, where pre-trained models are combined to generate samples from novel distributions.
Score-based diffusion models, on the other hand, are more widely adopted and come with a rich ecosystem of pre-trained models. However, they do not, in general, define an underlying energy function, making MH-based sampling inapplicable. In this work, we address this limitation by retaining score parameterization and introducing a novel MH-like acceptance rule based on line integration of the score function.
This allows the reuse of existing diffusion models while still combining the reverse process with various MCMC techniques, viewed as an instance of annealed MCMC.
Through experiments on synthetic and real-world data, we show that our MH-like samplers {yield relative improvements of similar magnitude to those observed} with energy-based models, without requiring explicit energy parameterization.
\end{abstract}

\section{Introduction}

Significant advancements have recently been achieved in generative modeling across various domains~\citep{brock2018large, brown2020language, ho2020denoising}.
These models have become potent priors for a wide range of applications, including code generation~\citep{li2022competition}, text-to-image generation~\citep{saharia2022photorealistic}, \mbox{question-answering ~\citep{brown2020language},} and many others~\citep{gungor2023adaptive, wynn2023diffusionerf}.
Among generative models, diffusion models~\citep{sohl2015deep, song2019generative, ho2020denoising} have arguably emerged as the most powerful class.
Diffusion models learn to denoise corrupted inputs in small, gradual steps and are capable of generating samples from complex distributions. They have been successful in many domains, such as generating highly realistic images~\citep{dhariwal2021diffbeatgan}, modeling temporal point processes~\citep{ludke2023diff_tpp}, and even generating neural network parameters~\citep{wang2024neural}.

Diffusion models also offer the capability of composed sampling, which combines pre-trained models to generate samples from a new distribution.
This approach, known as model composition, has a rich history~\citep{jacobs1991adaptive, hinton2002training, mayraz2000recognizing, liu2022compositional}.
For diffusion models, the most common form of composition is classifier-guided sampling, where the reverse process is augmented by a separate classifier model~\citep{sohl2015deep, dhariwal2021diffbeatgan, ho2021classifierfree}, but other compositions have also been explored~\citep{du2023reduce}.
The ability to compose new models without having to re-learn the individual components is especially appealing for diffusion models since their ever-increasing size and data hunger make them exceedingly costly to train~\citep{aghajanyan2023scaling}.
Therefore, developing sampling methods that work for pre-trained diffusion models is valuable.

The foundation of composed sampling for diffusion models is score-based, where we interpret diffusion models as predictors of the score function for the marginal distribution at each diffusion step~\citep{song2021scorebased}. From this perspective, MCMC methods, such as the Langevin algorithm (LA)~\citep{roberts2002mala} or Hamiltonian Monte Carlo (HMC) sampling~\citep{duane1987hmc}, emerge as viable options to incorporate. Augmenting the standard reverse process with additional MCMC sampling has been shown to improve composed sampling for diffusion models~\citep{du2023reduce, song2021scorebased}. However, we are {traditionally} restricted to unadjusted variants of these samplers, namely Unadjusted LA (U-LA) and Unadjusted HMC (U-HMC), which only require utilization of the score. 
{For instance, the seminal work by \citet{song2019generative} proposed score-based generative modeling by combining score networks with unadjusted Langevin dynamics. While highly effective, these unadjusted methods lack a rejection mechanism and may introduce discretization bias at larger step sizes. Correcting this bias requires evaluating the unnormalized density for a Metropolis--Hastings (MH) step~\citep{metropolis1953mh,hasting1970mh}—something that is not directly possible in standard score-parameterized diffusion models since they do not provide access to the underlying energy.}

An intriguing alternative to directly modeling the score function is to model the marginal distribution with an energy function, from which the score can be obtained through explicit differentiation~\citep{salimans2021should, song2019generative}. This parameterization connects diffusion models and energy-based models (EBMs)~\citep{lecun2006tutorial} and offers several desirable properties. With energy parameterization, we can evaluate the unnormalized density and guarantee a proper score function. This, in turn, enables an MH correction step when employing an MCMC method, where the MH acceptance probability is computed from the energy function. Adding such a correction step has been shown to improve sampling performance in composed models~\citep{du2023reduce}. Nevertheless, the score parameterization remains far more popular, as it avoids the direct computation of the gradient of the log density. {In practice, score-based models are significantly easier to train and scale to large datasets and high-dimensional problems than energy-based parameterizations, a practical advantage that has driven their widespread adoption.}

In this study, we build on the work in~\citep{du2023reduce} and introduce a novel approach to obtain an MH-like correction step directly from pre-trained diffusion models without relying on an energy-based parameterization. Specifically, we use a connection between the score and the energy to estimate the MH acceptance probability by approximating a line integral along the vector field generated by the score. This enables an improved sampling procedure for various pre-trained score-parameterized diffusion models. 
{We find that our approximate method yields relative improvements of a similar magnitude to those obtained with explicit energy parameterizations, without having to estimate the energy directly.}

In summary, our main contributions are:
\begin{itemize}
\item We show that MH-like correction sampling can be directly applied to score-based models without requiring additional training.
\item We introduce two efficient algorithms to approximate the energy difference used in MH and demonstrate that our pseudo-energy difference more accurately represents analytical energy differences than an explicitly trained energy model in a toy example while performing on par with the energy model on MNIST.
\item {We establish that the relative improvements in sample quality achieved via MCMC for energy-based models can be closely matched within a purely score-based framework.} In our implementations, the pseudo-MH correction also exhibits favorable practical runtime behavior compared to energy-based MCMC.
\end{itemize}

\section{Background}
\subsection{Diffusion Models}
We consider Gaussian diffusion models initially proposed by \cite{sohl2015deep} and further improved by \cite{song2019generative, ho2020denoising}.
Starting with a sample from the data distribution $\x_0 \sim q(\cdot)$, we construct a Markov chain of latent variables $\x_1,\ldots,\x_T$ by iteratively introducing Gaussian noise to the sample $q(\x_t|\x_{t-1}) = \normal{\x_t; \sqrt{1 - \beta_t} \x_{t-1}, \beta_t I}$,
where $\beta_t \in [0, 1), \, \forall t = 1, \dots, T$ are known.
For large enough $T$ we have $q(\x_T) \approx \normal{\x_T; 0, I}$.

A diffusion model learns to gradually denoise samples by modeling the distribution of the previous sample in the chain $\pEbm(\x_{t-1} \mid \x_t), t = 1, \dots, T$.
Approximate samples from the data distribution $q(\x_0)$ are obtained by starting from $\x_T \sim \normal{0, I}$ and sequentially sampling less noisy versions of the sample until the noise is removed.
This is called the \textit{reverse process}.

\textls[-18]{The reverse distribution is typically modeled as $p_\theta(\x_{t-1}|\x_t) = \mathcal{N}(\x_{t-1}; \mu_\theta(\x_t, t), \Sigma_\theta(\x_t, t))$,} since the posterior \( q(\x_{t-1}|\x_t) \) can be well approximated by a Gaussian distribution when the noise magnitude \( \beta_t \) is sufficiently small. The mean is parameterized as \( \mu_\theta(\x_t, t) = \frac{1}{\sqrt{\alpha_t}} \left( \x_t - \frac{\beta_t}{\sigma_t} \epsilon_\theta(\x_t, t) \right) \), where \( \alpha_t \) and $\sigma_t$ are positive and defined by \( \{ \beta_t \}_{t=1}^T \) \cite{ho2020denoising}. The noise prediction model \( \epsilon_\theta(\x_t, t) \), typically a neural network, is learned from data. We assume \( \Sigma_\theta(\x_t,t) = \beta_t I \) throughout unless otherwise stated.

\subsection{Energy-Based Models}

Energy-based models (EBMs) represent probability distributions with a scalar, non-negative energy function $\energy$
by assigning low energy to regions of the input space where the probability is high and high energy to regions where the distribution has little or no support:
\begin{equation}
\begin{aligned}
  \pEbm(\x_t, t) &= \frac{1}{\partFn(t)} \exp \left(-\dfrac{1}{\sigma_t} \energy(\x_t, t) \right), \\
  \partFn(t) &= \int \exp \left(- \dfrac{1}{\sigma_t}\energy(\x_t, t) \right) \dd \x_t.
\end{aligned}
\label{eq:ebm}
\end{equation}
{Here,} 
 we define $\energy$ as a time-dependent function and deliberately choose not to absorb $\sigma_t$ (introduced in the previous section) into $\energy$, to maintain a more explicit connection to diffusion models, as clarified in the next section.
This time dependency can be seen as a sequence of energy functions, one for each diffusion step $t$.
The normalization constant $\partFn$ is typically intractable, prohibiting computing a normalized density.
However, $\partFn$ does not depend on the input $\x_t$, making the so-called \textit{score function} easy to compute,
\begin{equation}
  \label{eq:ebm:score}
  \gradX \log \pEbm(\x_t, t) = - \dfrac{1}{\sigma_t} \gradX \energy(\x_t, t), 
\end{equation}
even though the gradient of the energy function can be costly to compute in practice.

\subsection{Energy and Score Parameterized Diffusion Models}

In the diffusion setting, one approach to training EBMs is denoising score matching (DSM). 
When the data are perturbed with Gaussian noise, the DSM loss coincides with the diffusion training loss (up to a factor of $\sigma_t^2$)~\citep{song2021scorebased}. 
This equivalence arises by identifying the noise prediction model, $\epsP(\x_t, t)$, with the score of an energy function, i.e., as an EBM:
\begin{align}
\label{eq:score_noise_connection}
    \epsP(\x_t, t) = \gradX \energy(\x_t, t).
\end{align}
Under this identification, $\epsP(\x_t, t)$ is required to define a proper score. 
Thus, an EBM and a plain diffusion model differ only in parameterization: the former uses \textit{energy parameterization} via $\energy$, while the latter, since $\epsP$ is only a pseudo-score, is referred to as using \mbox{\textit{score parameterization}.}

Both parameterizations have their advantages and disadvantages.
The energy parameterization can evaluate density $\pEbm(\x_t, t)$ up to normalization $\partFn(t)$, which enables various MCMC methods.
Furthermore, by making the score equal to the gradient of an actual scalar function, we ensure a proper score.
On the other hand, to evaluate the score function, $\energy$ must be explicitly differentiated, which can be costly.

The score parameterization is more flexible as it predicts an arbitrary vector field. While there is some empirical evidence that this improves sampling performance in diffusion processes~\citep{du2023reduce}, this difference may primarily stem from model architecture~\citep{salimans2021should}. Nevertheless, the score parameterization’s direct estimation of the score function makes it more efficient for reverse process sampling and remains the more widely adopted approach. In the next section, we describe how these parameterizations affect the design of MCMC samplers for diffusion models.

\subsection{MCMC Sampling for Diffusion Models}

MCMC sampling is a promising strategy for improving diffusion model sampling since it can be combined with the reverse process.
MCMC methods are naturally defined through transition kernels, and just like the reverse process there are MCMC methods that base their transitions on the score function, such as the Unadjusted Langevin Algorithm (U\!-LA) and the Unadjusted Hamiltonian Monte Carlo (U\!-HMC)~\citep{roberts1996exponential,geffner2021mcmc}.
We let $\tau$ denote the index of the MCMC iterations (as opposed to $t$, which refers to the diffusion timestep).

For U\!-LA we use the kernel
\begin{align*}
    \mcKern{\x^{\tau+1} \mid \x^{\tau}}
    \;=\;
    \normal{\x^{\tau+1};\; \x^{\tau} + \langStep_t \gradX \log \pEbm(\x^{\tau}, t),\; 2 \langStep_t I},
\end{align*}
at diffusion step $t$, where $\x^0 = \x_t$, $\langStep_t$ is the step size, and the chain is iterated for $\mcSteps$ steps.

For U\!-HMC we augment the state with momenta $\vek{v}^\tau \sim \normal{0, M_t}$ (diagonal mass $M_t$), and propose a new state by applying $\ell_t$ leapfrog steps of size $\varepsilon_t$ under the same score field. 
Writing
\vspace{-6pt}
\begin{align*}
(\x^{\tau+1}, \vek{v}^{\tau+1}) \;=\; \mathrm{LF}_{\ell_t,\varepsilon_t}\!\big(\x^\tau,\vek{v}^\tau\big),
\end{align*}
for the leapfrog map (which in our case is defined with the score $\nabla_x \log \pEbm(\x,t)$ as potential gradient), 
the kernel for $\x$ is given implicitly by this deterministic proposal after marginalizing out $\mathbf{v}$.
In practice, we may perform $L_t$ such proposals per diffusion step $t$.

These methods are called unadjusted since, as $\mcSteps$ grows, the Markov chains converges to the target distribution only in the limit of infinitesimal step sizes.
By adding, for instance, a Metropolis--Hastings (MH) correction step, we can sample with larger step sizes and still converge to the target distribution~\citep{metropolis1953mh,hasting1970mh}.
With the correction, we sample a candidate $\xCand \sim \mcKern{\cdot \mid \x^\tau}$ and accept it as the new iterate with probability
\begin{align}
    \label{eq:mh:acc_prob}
    \acc \;=\; \min \!\left( 1,\;
    \frac{\pEbm(\xCand, t)}{\pEbm(\x^\tau, t)}\;
    \frac{\mcKern{\x^\tau \mid \xCand}}{\mcKern{\xCand \mid \x^\tau}}
    \right),
\end{align}
so that $\x^{\tau+1}=\xCand$ with probability $\acc$, and $\x^{\tau+1}=\x^\tau$ otherwise. With the correction, U\!-LA becomes LA (often referred to as MALA)~\citep{besag1994mala}, and U\!-HMC becomes the standard HMC algorithm~\citep{duane1987hmc,neal1996bayes}.

The model $\pEbm$ appears only through a ratio, so a normalized density is not required.
When $\pEbm$ is parameterized as an EBM (see~(\ref{eq:ebm})), the ratio simplifies to
\begin{align}
    \label{eq:mh:rel_energy}
    \frac{\pEbm(\xCand, t)}{\pEbm(\x^\tau, t)}
    \;=\;
    \exp\!\left(\frac{1}{\sigma_t}\big(\energy(\x^\tau, t) - \energy(\xCand, t)\big)\right),
\end{align}
which allows us to directly evaluate the MH acceptance probability, making it straightforward to construct an adjusted MCMC sampler. This offers a key advantage over the score parameterization, where only an approximation of the score is accessible which cannot directly be used to compute the probability ratio needed in MH.

\subsection{Sampling from Composed Models}

Composed sampling is a powerful feature of diffusion models that enables sampling from new target distributions by combining multiple pre-trained models. Rather than retraining a model for every new task or data combination, one can reuse existing components. This flexibility is especially appealing in large-scale settings, where retraining is often prohibitively expensive.

The most common form of composition is \textit{guidance}~\citep{dhariwal2021diffbeatgan}, where the goal is to sample from a distribution conditioned on a class label $y$,
\begin{align}
\label{eq:guidance:post}
    q(\x_0 \mid y) \propto q(\x_0) q(y \mid \x_0).
\end{align}
This is implemented by modifying the score function at each diffusion step as
\begin{align}
\label{eq:guidance:score}
    \gradX \log \pEbm(\x_t, t) + \lambda \gradX \log \pCfull (y \mid \x_t, t),
\end{align}
where $\pEbm$ is an unconditional diffusion model and $\pCfull$ is a classifier predicting class $y$. Hyperparameter $\lambda$ controls the strength of the conditioning. We refer to this approach as \textit{classifier-full guidance}. Other variants include reconstruction guidance~\citep{chung2023reconstrguidance, ho2022guidance} and classifier-free guidance~\citep{ho2021classifierfree}.

More generally, \cite{du2023reduce} explores a range of composition types beyond guidance, including \textit{products}, \textit{negations}, and \textit{mixtures}. A product composition—of which guidance can be seen as a special case—is defined as
\begin{equation}
\label{eq:comp:prod}
    \qProd(\x_0) \propto \prod_i q^i(\x_0),
\end{equation}
and leads to the composed model at diffusion step $t$,
\begin{equation}
\label{eq:comp:prod:ebm}
\begin{aligned}
    \pProd(\x_t, t) \propto \prod_i \pEbmi^i(\x_t, t)
    = \exp \left( -\frac{1}{\sigma_t} \sum_i \energyi^i(\x_t, t) \right).
\end{aligned}
\end{equation}
This distribution is then used as the target in MCMC sampling, resulting in improved sampling performance.

Importantly, the factorization in \eqref{eq:comp:prod} only strictly holds at $t = 0$; at intermediate diffusion steps, the composed model $\pProd(\x_t, t)$ does not generally correspond to the true marginal of any product data distribution; refer to~\citet{du2023reduce}. 
This becomes problematic when relying solely on the reverse process, which assumes access to a valid score function for the true intermediate marginals. However, this construction remains valid and effective from the perspective of \textit{annealed MCMC}~\cite{neal2001annealed}, where the overall sampling procedure is interpreted as a chain targeting a sequence of gradually evolving distributions. From this viewpoint, the intermediate distributions $\pProd(\x_t, t)$ are treated as design choices that guide the chain toward the final target $\qProd(\x_0)$, and asymptotic correctness is still preserved. In practice, since diffusion models are trained using denoising score matching, the sampling process converges to a denoised version of $\qProd(\x_0)$, which can be made arbitrarily close to the true distribution by construction.

Note that for models using a score-based parameterization, a pseudo-score for this type of composition is equal to $- \frac{1}{\sigma_t} \sum_i \epsPi^i(\x_t, t)$.
\section{MCMC Correction Step for Score Parameterization}

We propose combining the energy parameterization properties with the performance and practical accessibility of score parameterization. Instead of using energy parameterization and computing the score by differentiation, we take the complementary approach: using score parameterization and computing the change in (pseudo-)energy by integrating the score.

\subsection{Pseudo-Energy Difference and MH-like Correction}

This section describes how MCMC acceptance probabilities can be approximated given
only access to a score function.
The Metropolis--Hastings acceptance probability \mbox{in
(\ref{eq:mh:acc_prob})} depends on the \emph{relative} probability of a proposed
state $\xCand$ compared to the current state $\x^\tau$.
Since the proposal kernel $\mcKern{\cdot\mid\cdot}$ is straightforward to
evaluate, the acceptance ratio reduces to comparing unnormalized target
densities.  

For an energy-based model, this ratio depends only on the difference and—because $E_\theta$ is a scalar potential—this difference can be written as a
path-independent line integral,
\begin{align}
    E_\theta(\x^\tau,t) - E_\theta(\xCand,t)
    &= - \!\int_{\mathcal{C}}
        \nabla_\xi E_\theta(\xi,t)\cdot d\xi,
    \label{eq:line_integral_energy}
\end{align}
for any differentiable curve $\mathcal{C}$ connecting $\x^\tau$ and $\xCand$.
This provides a natural interpretation of MH in terms of integrating the model's
score field along a path.

For score-parameterized diffusion models, the situation is reversed:
the model provides a score field $\epsilon_\theta(x,t)$ but no explicit energy
function.  
Motivated by the fact that denoising score matching trains $\epsilon_\theta$ to
approximate the gradient of the log-density under Gaussian perturbations, we
define an analogous \emph{pseudo-energy difference}
\begin{align}
\label{eq:pseudo_energy_diff}
    \Delta\tilde{E}_{\mathcal{C}}(x \!\to\! x',t)
    = - \!\int_{\mathcal{C}} \epsilon_\theta(\xi,t)\cdot d\xi,
\end{align}
where $\mathcal{C}$ is any smooth curve connecting $x$ and $x'$.
This construction can be viewed as integrating the vector field
$\epsilon_\theta$ along a path, thereby approximating the change in an
underlying scalar potential---if such a potential existed.

Based on this quantity, we define an MH-like acceptance probability
\begin{align}
\label{eq:mh_like}
\alpha_{\mathrm{MH\text{-}like}}
    =
    \min\!\left(
    1,\;
    \exp\!\left[\frac{1}{\sigma_t}\Delta\tilde{E}_{\mathcal{C}}(x\!\to\!x',t)\right]
    \frac{k_t(x \mid x')}{k_t(x' \mid x)}
    \right).
\end{align}
If $\epsilon_\theta(x,t)=\nabla_x F_\theta(x,t)$ for some potential $F_\theta$, 
then $\Delta \tilde{E}_C$ equals $F_\theta(x',t)-F_\theta(x,t)$ for all curves $C$, 
and (\ref{eq:mh_like}) exactly recovers the true MH acceptance probability by the fundamental theorem of line integrals.

In practice, $\epsilon_\theta$ is not perfectly conservative and the pseudo-energy 
difference depends on the chosen path $\mathcal{C}$.  
We therefore consider two practical choices: 
(i)~a straight line between $x$ and $x'$ and 
(ii)~a curved trajectory following the leapfrog path of an HMC proposal.  
The latter allows reusing score evaluations already computed during proposal generation, 
yielding higher numerical accuracy without additional model evaluations.

In both cases, the integral in (\ref{eq:pseudo_energy_diff}) is approximated using 
the trapezoidal rule, with the number of line segments treated as a hyperparameter $n$.  
This requires score evaluations at internal points of $\mathcal{C}$, but avoids 
differentiating the model; by contrast, the energy parameterization evaluates 
$E_\theta$ at the endpoints but must differentiate it to obtain the score.

An overview of the full sampling procedure is given in 
Algorithm~\ref{alg:annealed_mcmc}.
At each diffusion step, an optional reverse update is followed by an MCMC refinement 
targeting the intermediate distribution.  
This structure aligns naturally with the annealed MCMC framework, where both the 
reverse update and the MCMC kernel act as design choices guiding the chain toward 
the final distribution.  
Including the reverse step typically improves sample quality~\citep{du2023reduce}.
%1
\begin{algorithm}[H]
\caption{Annealed MCMC with MH-like correction}
\label{alg:annealed_mcmc}
\begin{algorithmic}[1]
\Require Score function $\epsilon_\theta(\cdot,t)$; reverse-diffusion schedule $(\alpha_t,\beta_t,\sigma_t)_{t=1}^T$; kernel family $\{k_t(\cdot\mid\cdot)\}_{t=1}^T$ (LA or HMC) with kernel hyperparameters; 
 MCMC steps per diffusion step $\{L_t\}$; integration mode
 $\texttt{mode}\!\in\!\{\texttt{line},\texttt{curve}\}$; line segments $n$ (if \texttt{line}); per-leapfrog subsegments $m$ (if \texttt{curve}).
\State Sample initial $x_T \sim \mathcal{N}(0,I)$.
\For{$t = T$ \textbf{to} $1$}
  \State \textit{(Optional) Reverse step:} update $x_{t-1}$ from $x_t$ using the standard reverse-diffusion update at time $t$
  \If{$t > 1$}
    \State $t' \gets t-1$ \Comment{all MCMC quantities live at time $t'$}
    \State Initialize MCMC: $x^{0} \gets x_{t'}$.
    \For{$\tau = 1$ \textbf{to} $L_t$}
        \State Propose candidate $\hat{x} \sim k_{t'}(\cdot \mid x^{\tau-1})$ \Comment{LA or HMC kernel}
        \State $\Delta\tilde{E}_{\mathcal{C}} \gets \texttt{EstimatePseudoEnergy}(x^{\tau-1}, \hat{x}, t';\ \texttt{mode}, n, m)$ 
    \Comment{Algorithm~\ref{alg:line} or Algorithm~\ref{alg:curve}}
        \State Compute acceptance\vspace{-6pt}
        \[
        \alpha_{\mathrm{MH\text{-}like}} \;=\; \min\!\left(1,\ \exp\!\Big(\tfrac{1}{\sigma_{t'}}\,\Delta\tilde{E}_{\mathcal{C}}\Big)\cdot \frac{k_{t'}(x^{\tau-1}\mid \hat{x})}{k_{t'}(\hat{x}\mid x^{\tau-1})} \right)
        \]
        \State With probability $\alpha_{\mathrm{MH\text{-}like}}$: $x^\tau \gets \hat{x}$; otherwise $x^\tau \gets x^{\tau-1}$.
    \EndFor
  \State Set $x_{t'} \gets x^{L_t}$.
  \EndIf
\EndFor
\State \Return $x_0$
\end{algorithmic}
\end{algorithm}

\vspace{-6pt}

%2
\begin{algorithm}[H]
\caption{Estimate Pseudo-Energy (straight-line path)}
\label{alg:line}
\begin{algorithmic}[1]
\Require Current $x$, candidate $\hat{x}$, time $t$, number of segments $n \ge 2$; access to $\epsilon_\theta(\cdot,t)$. Note that $\epsilon_\theta(x,t)$ and $\epsilon_\theta(\hat{x},t)$ are available from cache (Algorithm~\ref{alg:annealed_mcmc}). The routine may reuse them instead of re-evaluating
\State $\Delta r \gets \tfrac{1}{n-1}\,(\hat{x}-x)$
\State $r_{\mathrm{prev}} \gets x$;\quad $g_{\mathrm{prev}} \gets \epsilon_\theta(x,t)$;\quad $\Delta\tilde{E}_{\mathcal{C}} \gets 0$
\For{$j=1$ \textbf{to} $n-1$}
   \State $r_j \gets r_{\mathrm{prev}} + \Delta r$
   \State $g_j \gets \epsilon_\theta(r_j,t)$ \Comment{reuse cached value at endpoints}
   \State $\Delta\tilde{E}_{\mathcal{C}} \gets \Delta\tilde{E}_{\mathcal{C}} - \tfrac{1}{2}\,\big(g_{\mathrm{prev}} + g_j\big)\cdot \Delta r$
   \State $r_{\mathrm{prev}} \gets r_j$;\quad $g_{\mathrm{prev}} \gets g_j$
\EndFor
\State \Return $\Delta\tilde{E}_{\mathcal{C}}$
\end{algorithmic}
\noindent\textit{Complexity (extra score evaluations).} If endpoint scores are reused, this routine performs $n-2$ new $\epsilon_\theta(\cdot,t)$ calls; otherwise $n$ calls.
\end{algorithm}
\unskip
%3
\begin{algorithm}[H]
\caption{Estimate Pseudo-Energy (HMC-curved path)}
\label{alg:curve}
\begin{algorithmic}[1]
\Require Current $x$, candidate $\hat{x}$, time $t$; leapfrog trajectory $\{x^{(i)}\}_{i=0}^{\ell_t}$ with $x^{(0)}{=}x$, $x^{(\ell_t)}{=}\hat{x}$; subsegments per leapfrog step $m \ge 2$; access to $\epsilon_\theta(\cdot,t)$. Note that $\{\epsilon_\theta(x^{(i)},t)\}_{i=0}^{\ell_t}$ are available from cache (Algorithm~\ref{alg:annealed_mcmc}) and may be reused instead of re-evaluating.
\State $\Delta\tilde{E}_{\mathcal{C}} \gets 0$
\For{$i=1$ \textbf{to} $\ell_t$} \Comment{integrate along each LF segment $x^{(i-1)} \to x^{(i)}$}
   \State $r_{\mathrm{prev}} \gets x^{(i-1)}$;\quad $g_{\mathrm{prev}} \gets \epsilon_\theta(x^{(i-1)},t)$
   \State $\Delta r_i \gets \tfrac{1}{\,m-1\,}\big(x^{(i)} - x^{(i-1)}\big)$
   \For{$j=1$ \textbf{to} $m-1$}
      \State $r_{i,j} \gets r_{\mathrm{prev}} + \Delta r_i$
      \State $g_{i,j} \gets \epsilon_\theta(r_{i,j},t)$ \Comment{reuse cached leapfrog scores if available}
      \State $\Delta\tilde{E}_{\mathcal{C}} \gets \Delta\tilde{E}_{\mathcal{C}} - \tfrac{1}{2}\,\big(g_{\mathrm{prev}} + g_{i,j}\big)\cdot \big(r_{i,j} - r_{\mathrm{prev}}\big)$
      \State $r_{\mathrm{prev}} \gets r_{i,j}$;\quad $g_{\mathrm{prev}} \gets g_{i,j}$
   \EndFor
\EndFor
\State \Return $\Delta\tilde{E}_{\mathcal{C}}$
\end{algorithmic}
\noindent\textit{Complexity.} With cached leapfrog scores, this routine performs $\ell_t\,(m-2)$ new $\epsilon_\theta(\cdot,t)$ calls (one per interior subsegment); without caching it performs $\ell_t\,(m-1)$ calls. Setting $m=2$ reduces to a trapezoid rule that uses only leapfrog endpoints.
\end{algorithm}

\subsection{MH Correction for Composition Models}
The pseudo-energy difference for compositions can be derived based on their specific definitions. Our proposed method applies directly to product compositions. We calculate a pseudo-energy difference, corresponding to $\eProd(\x^\tau, t) - \eProd(\xCand, t)$ for an EBM (defined in (\ref{eq:comp:prod:ebm})), as 
\begin{equation} 
\label{eq:energy_diff:prod} 
- \int_{\mathcal{C}} \sum_i \epsPi^i(\xi, t) \cdot d\xi.
\end{equation} 
Guidance is a specific case of product composition, where the pseudo-score is composed of two terms according to (\ref{eq:guidance:score}): the unconditional diffusion model $\epsP(\x_t, t)$ and the score of a classifier $\pCfull(y \mid \x_t, t)$. Since $\pCfull(y \mid \x_t, t)$ can be evaluated directly, only the pseudo-energy difference for $\epsP(\x_t, t)$ requires computation using the line integral in (\ref{eq:energy_diff:prod}).

The pseudo-energy difference for a negation composition (as defined in~\citep{du2023reduce}) can be computed analogously to products, as negations follow a similar additive structure in their pseudo-scores.

Mixture compositions (as defined in~\citep{du2023reduce}), on the other hand, cannot be expressed as a pseudo-energy difference, since mixtures do not naturally conform to an additive structure analogous to products or negations. However, mixtures can be addressed by first sampling a component distribution according to the mixture definition and then generating a sample from that distribution. The MH correction can subsequently be applied to this sampled distribution, providing a seamless way to handle mixture compositions within our framework.

This generalization allows our method to support advanced use cases such as classifier guidance, multi-modal fusion, and spatially structured prompts, without requiring retraining or access to energy-based models.
\section{Results}

In this section, we present an empirical evaluation of our MH-like correction method\footnote{Code available at \url{https://github.com/FraunhoferChalmersCentre/mcmc_corr_score_diffusion}}, 
examining both the accuracy of the pseudo-energy differences and the quality of the generated samples. The experiments are designed to span a spectrum of difficulty: from controlled, low-dimensional setups where models can be trained from scratch and analytical solutions are available to more realistic high-dimensional scenarios involving pre-trained models. Our two primary objectives are (1) to compare our proposed approach against a true energy parameterization when available and (2) to assess the sampling improvements achieved over the standard reverse process when augmented with MCMC steps.

The experiments in Sections~\ref{section:pseudo-energy} and \ref{section:2dcomp} and the first part of Section \ref{section:guideddiffusion} involve training diffusion models using both energy and score parameterizations. The score parameterization follows a noise prediction model, $\epsP(\x_t, t)$, while the energy parameterization defines an energy function as $\energy(\x_t, t) = \twoNorm{\x_t - \score(\x_t, t)}^2$, as in~\citep{du2023reduce}. We use identical network architectures for $\epsilon_\theta$ and $s_\theta$. Both models are trained with the standard diffusion loss~\citep{ho2020denoising}, with the energy model’s score function obtained through explicit differentiation.

The later experiments utilize only pre-trained score-based diffusion models, as pre-trained energy-based models are unavailable for direct comparison. We evaluate both unadjusted and MH-corrected versions of Langevin and Hamiltonian Monte Carlo, comparing them against the standard reverse process, which serves as the baseline.

For the MH-like correction, we examine two types of integration paths: 
a straight line between $\x^\tau$ and $\xCand$ (Algorithm~\ref{alg:line}) 
and the trajectory defined by the HMC leapfrog steps (Algorithm~\ref{alg:curve}). 
Both approaches rely on a trapezoidal rule where the number of intermediate points 
is treated as a hyperparameter.

Complete training details, hyperparameter settings, and implementation specifics are deferred to Appendix~\ref{sec:expdetails}.
\subsection{Evaluating Pseudo-Energy Differences} \label{section:pseudo-energy}

To evaluate the accuracy of pseudo-energy differences, we conducted experiments on a synthetic 2D dataset, generated from a bivariate Gaussian distribution to allow access to analytical solutions, and a higher-dimensional dataset, MNIST~\citep{mnist}. For each experiment, we trained 10 independent score models and 10 independent energy models from scratch. For evaluation, we sampled 2000 pairs of points $(x_t^1, x_t^2)$ independently via the forward process at various diffusion steps $t$. {These pairs were used to compute the score-based pseudo-energy difference ($\Delta\tilde{E}_{\text{score}}$) and the explicit EBM energy difference ($\Delta E_{\text{EBM}}$), as well as the analytical difference ($\Delta E_{\text{true}}$) when available.} The pseudo-energy difference was computed along a straight-line path connecting the two points, using five discretization points for the numerical integration.

\textbf{2D Gaussian:}
For the 2D Gaussian dataset, the relative error metric is defined as 
$|\Delta E_{\text{pred}} - \Delta E_{\text{true}}| / |\Delta E_{\text{true}}|$, 
where $\Delta E_{\text{pred}}$ {corresponds to either the explicitly predicted difference from the energy model ($\Delta E_{\text{EBM}}$) or the pseudo-energy difference from the score model ($\Delta\tilde{E}_{\text{score}}$)}, and $\Delta E_{\text{true}}$ is the analytical energy difference. The median relative error was calculated across all sampled pairs for each trained model, and the mean and standard deviation of this metric were computed across the 10 models. Interestingly, the score model achieved a lower relative error ($0.071 \pm 0.005$) compared to the energy model ($0.084 \pm 0.004$), demonstrating {that the line-integral approximation aligns slightly better} with the true energy differences {in this setting}.

\textbf{MNIST:}  
For the MNIST dataset, where analytical energy differences are unavailable, we used a symmetric relative {discrepancy} metric defined as 
{$2 |\Delta\tilde{E}_{\text{score}} - \Delta E_{\text{EBM}}| / (|\Delta\tilde{E}_{\text{score}}| + |\Delta E_{\text{EBM}}|)$}. 
The median relative {discrepancy} was calculated across all sampled pairs for each trained model, and the mean and standard deviation were computed across the \mbox{10 models.} This yielded a mean symmetric relative discrepancy of $0.030 \pm 0.002$, indicating that the {pseudo-energy} differences predicted by the score models align closely {with the explicit EBM predictions}, even in this higher-dimensional setting.

\subsection{Two-Dimensional Composition} \label{section:2dcomp} 

To investigate the effectiveness of our MH-like correction in a controlled yet expressive setting, we replicate the 2D composition experiment introduced by~\citet{du2023reduce}. We use their publicly available implementation\footnote{Code available at \url{https://github.com/yilundu/reduce_reuse_recycle}} as a foundation. Only minor modifications are necessary, ensuring a faithful reproduction of their setup. Apart from differences in evaluation metrics, our setup is identical to theirs.

A 2D density pair is composed via multiplication into a complex distribution, as \mbox{in (\ref{eq:comp:prod:ebm}):} a Gaussian mixture with eight modes in a circle and a uniform distribution covering two of the modes. For a visual representation
of the two individual distributions and their resulting product distribution together with samples from the reverse diffusion and HMC corrected samples, see Figure \ref{fig:exp:2d_toy}. The baseline reverse diffusion process uses $T=100$ steps. In the MCMC variants, following \cite{du2023reduce}, we omit the optional reverse step for a fair comparison. MCMC sampling runs for $\mcSteps = 10$ at each $t$, with (U-)HMC using three leapfrog steps per MCMC step.
\vspace{-6pt}
\begin{figure}[H]

    \resizebox{1\linewidth}{!}{%
   \begin{tikzpicture}[
    font=\small,
    scale=1.0
]
\def\markSize{1pt}
\begin{groupplot}[
    group style={
        group size=6 by 1,
        horizontal sep=0.25cm,
        vertical sep = 2cm,
    },
    height=3.625cm,
    width= 3.625cm,
    xlabel={},
    xmin=-0.75,
    xmax=0.75,
    ylabel={},
    ymin=-1,
    xtick={-0.5, 0.0, 0.5},
    ytick={-1, 0.0, 1},
    ymax=1,
    tick align=outside,
    tick pos=left,
]
% Gauss mix (GMM)
\nextgroupplot[
title={(\textbf{a}) Gaussian \\ mixture},
% Necessary for linebreak in the center
align = center,
mark size=\markSize,
]
  \addplot[blue, only marks]
  table[
      x=x,
      y=y,
  ]
  {figures/two_d_comp/gmm.txt};
%
% Uniform
\nextgroupplot[
title={(\textbf{b}) Uniform},
% Necessary for linebreak in the center
align = center,
mark size=\markSize,
ymajorticks=false,
]
  \addplot[blue, only marks]
  table[
      x=x,
      y=y,
  ]
  {figures/two_d_comp/bar.txt};
% Product
\nextgroupplot[
title={(\textbf{c}) Product},
% Necessary for linebreak in the center
align = center,
mark size=\markSize,
ymajorticks=false,
]
  \addplot[blue, only marks]
  table[
      x=x,
      y=y,
  ]
  {figures/two_d_comp/prod.txt};
% Reverse
\nextgroupplot[
title={(\textbf{d}) Reverse},
% Necessary for linebreak in the center
align = center,
mark size=\markSize,
ymajorticks=false,
]
  \addplot[blue, only marks]
  table[
      x=x,
      y=y,
  ]
  {figures/two_d_comp/score_reverse.txt};
% HMC score
\nextgroupplot[
title={(\textbf{e}) HMC \\ Score},
% Necessary for linebreak in the center
align = center,
mark size=\markSize,
ymajorticks=false,
]
  \addplot[blue, only marks]
  table[
      x=x,
      y=y,
  ]
  {figures/two_d_comp/score_hmc.txt};
% HMC Energy
\nextgroupplot[
title={(\textbf{f}) HMC \\ Energy},
% Necessary for linebreak in the center
align = center,
mark size=\markSize,
ymajorticks=false,
]
  \addplot[blue, only marks]
  table[
      x=x,
      y=y,
  ]
  {figures/two_d_comp/energy_hmc.txt};
\end{groupplot}
\end{tikzpicture}

    }
    \caption{Samples
 from:
    {(\textbf{a},\textbf{b})} the component distributions: a Gaussian mixture and a uniform distribution,
   {(\textbf{c})} the true product distribution,
   {(\textbf{d})} a standard score-parameterized reverse process,
   ({\textbf{e},\textbf{f})} HMC sampling using score and energy parameterization, respectively.
    }
    \label{fig:exp:2d_toy}
\end{figure}

We evaluate performance using three metrics:  
(1) negative log-likelihood (NLL),  
\mbox{(2) a Gaussian} mixture model (GMM) comparison, and  
(3) the Wasserstein-2 distance ($W_2$).  
The suffixes in Table~\ref{tab:2d_perf} refer to the choice of integration path and the number of evaluation points:  
“L’’ denotes the straight-line path from Algorithm~\ref{alg:line}, while “C’’ denotes the curve path from Algorithm~\ref{alg:curve}.  
The number indicates trapezoidal evaluation points ($m$ or $n$).

\begin{table}[H]
\centering
\caption{
Performance metrics (NLL, GMM, and $W_2$) for the 2D composition experiment. The downward arrow ($\downarrow$) indicates that lower values are better. Values represent the mean with the standard deviation in parentheses over 10 independent trials. Bold numbers indicate the best performance in each metric for both Energy and Score samplers.}
\label{tab:2d_perf}
\begin{tabularx}{\textwidth}{CCCCC}
\toprule
 & \textbf{Sampler} & \textbf{NLL} \boldmath{$\downarrow$} 
 & \textbf{GMM}~\boldmath{$\downarrow$} & \boldmath{$W_2$~$\downarrow$} \\
\toprule
% --- Energy
\multirow{5}{*}{\rotatebox[origin=c]{0}{Energy}} %
& Reverse   & $8.22\,(0.21)$ 
 & $27.01\,(1.34)$ & $5.81\,(0.19)$ \\
& U-LA      & $7.52\,(0.22)$ & $14.61\,(1.35)$ & $4.19\,(0.45)$ \\
& LA        & $6.50\,(0.30)$ & $14.66\,(1.46)$ & $4.24\,(0.55)$ \\
& U-HMC     & $5.72\,(0.18)$ & $6.53\,(0.91)$  & $4.19\,(1.25)$ \\
& HMC       & \boldmath{$4.09\,(0.14)$}
 & \boldmath{${3.33\,(0.65)}$} & \boldmath{${4.12\,(1.44)}$} \\
\midrule
% --- Score
\multirow{10}{*}{\rotatebox[origin=c]{0}{Score}} 
& Reverse   & $8.15\,(0.24)$ & $26.88\,(1.20)$ & $5.80\,(0.20)$ \\
& U-LA      & $7.57\,(0.12)$ & $14.99\,(0.62)$ & $4.44\,(0.63)$ \\
& LA-3L     & $6.45\,(0.20)$ & $14.28\,(1.07)$ & $4.03\,(0.52)$ \\
& LA-5L     & $6.61\,(0.17)$ & $15.19\,(0.92)$ & $4.22\,(0.46)$ \\
& LA-10L     & $6.53\,(0.17)$ & $14.75\,(0.91)$ & $4.20\,(0.51)$ \\
& U-HMC     & $5.77\,(0.12)$ & $6.90\,(0.71)$  & $3.39\,(0.77)$ \\
& HMC-3L    & $4.29\,(0.13)$ & $3.72\,(0.61)$  & $2.92\,(1.02)$ \\
& HMC-5L    & \boldmath{${4.07\,(0.13)}$} & $3.08\,(0.69)$ & \boldmath{${2.68\,(1.20)}$} \\
& HMC-10L    & \boldmath{${4.07\,(0.14)}$} & $3.17\,(0.56)$ & $2.87\,(0.89)$ \\
& HMC-2C     & \boldmath{${4.07\,(0.12)}$} & \boldmath{${3.06\,(0.54)}$} & $2.94\,(0.90)$ \\
\bottomrule
\end{tabularx}
\end{table}

Table~\ref{tab:2d_perf} reports performance metrics averaged over 10 independent trials. In each trial, we train the diffusion models from scratch and sample 2000 points using different MCMC methods. The corrected sampling methods consistently outperform the unadjusted ones. HMC variants yield the best results across all metrics. Score- and energy-parameterized samplers show similar NLL and GMM performance, while HMC with score parameterization achieves a substantially lower $W_2$. Performance also saturates with as few as three integration points.

In addition to performance, we measure runtime and memory consumption.  
These results are reported separately in Table~\ref{tab:2d_runtime}.  
The experiment is implemented in JAX (v0.4.30)
~\citep{jax2018github} and run on a desktop computer equipped with an NVIDIA GeForce RTX 3060 GPU.  
Score-based parameterization is more than twice as memory-efficient as energy-based parameterization, and—except for LA with $m=10$—also faster for corresponding MCMC methods. Notably, the HMC curve variant is significantly faster.  
Although score-based corrections require more model evaluations, they do not necessarily incur higher runtime or memory costs.

\begin{table}[H]
%\centering
\caption{Runtime (seconds) and peak memory usage (MiB) for the 2D composition experiment.}
\label{tab:2d_runtime}
\begin{tabularx}{\textwidth}{CCCC}
\toprule
 & \textbf{Sampler} & \textbf{Time} & \textbf{Memory} \\
\midrule
% --- Energy
\multirow{5}{*}{\rotatebox[origin=c]{0}{Energy}} 
& Reverse   & $0.22(0.00)$ & $5252$ \\
& U-LA      & $1.54(0.01)$ & $5252$ \\
& LA        & $9.13(0.08)$ & $5252$ \\
& U-HMC     & $2.36(0.05)$ & $5254$ \\
& HMC       & $21.02(0.06)$ & $5256$ \\
\midrule
% --- Score
\multirow{10}{*}{\rotatebox[origin=c]{0}{Score}} 
& Reverse   & $0.11\,(0.00)$ & $2178$ \\
& U-LA      & $0.93\,(0.01)$ & $2180$ \\
& LA-3L     & $4.77\,(0.13)$ & $2180$ \\
& LA-5L     & $7.60\,(0.02)$ & $2180$ \\
& LA-10L     & $10.53\,(0.15)$ & $2180$ \\
& U-HMC     & $1.19\,(0.01)$ & $2180$ \\
& HMC-3L    & $7.08\,(0.02)$ & $2180$ \\
& HMC-5L    & $9.56\,(0.01)$ & $2180$ \\
& HMC-10L    & $10.64\,(0.01)$ & $2180$ \\
& HMC-2C     & $1.48 \,(0.04)$ & $2180$ \\
\bottomrule
\end{tabularx}
\end{table}

As discussed by~\citet{du2023reduce}, directly adding score functions does not yield a valid product composition, which explains why the reverse sampler performs poorly in this setting. This apparent failure is expected and highlights the motivation for annealed MCMC: by treating intermediate distributions as design choices that guide the chain toward the target, annealing achieves improved results while preserving asymptotic correctness. This accounts for the large gap between the reverse method and the annealed MCMC variants reported in Table~\ref{tab:2d_perf}.

\subsection{Guided Diffusion} \label{section:guideddiffusion}

We evaluate our proposed sampling methods for guided diffusion on the CIFAR-100~\citep{krizhevsky2009cifar} and ImageNet~\citep{deng2009imagenet} datasets. The sampling process is based on a score function defined in (\ref{eq:guidance:score}). For both datasets, the marginal score, $\gradX \log q(\x_t)$, is estimated using an unconditional diffusion model parameterized by a UNet architecture. For the guidance model, we use classifier-full guidance, training a time-dependent classifier to predict class labels across all diffusion steps, $\pCfull(y \mid \x_t, t)$. This classifier shares its architecture with the encoder part of the UNet used for the diffusion model and is extended with a dense output layer. The guidance scale is set to $\lambda = 20.0$ across all experiments, which is a commonly used default in classifier-guided diffusion models. This choice is also consistent with the settings provided in the public implementation\footnote{Code available at \url{https://github.com/yilundu/reduce_reuse_recycle}} of \cite{du2023reduce}.
Sampling is based on the standard reverse process with $T=1000$, and additional MCMC steps are incorporated to refine the generated samples. The \emph{Reverse} baseline reported in the tables corresponds to the standard classifier-full guided reverse diffusion process without any additional MCMC refinement steps.

To quantify generation quality, we use three evaluation metrics: the Fréchet Inception Distance (FID) \citep{heusel2017gans}, which compares the distribution of generated and real images; classification accuracy, based on a separate pre-trained classifier applied to generated samples; and, for ImageNet, an additional top 5 accuracy metric.

\textbf{CIFAR-100:} 
For CIFAR-100, we trained the diffusion models from scratch using the same UNet architecture and training settings as in \cite{ho2020denoising}, which were originally designed for CIFAR-10~\citep{krizhevsky2009cifar}. The MCMC samplers add $\mcSteps = 2$ or $6$ extra MCMC steps at each diffusion step $t$ for (U-)HMC and (U-)LA, respectively, with (U-)HMC using three leapfrog steps per MCMC step, following the configuration used by~\citet{du2023reduce} in their guided \mbox{diffusion experiments.}

For this experiment, we used denser meshes in the trapezoidal rule compared to the 2D setting. 
For HMC, we followed the curve defined by the leapfrog steps (Algorithm~\ref{alg:curve}) and set $n=3$, 
i.e., the leapfrog points plus an additional midpoint evaluation. 
For LA, we followed the straight line path (Algorithm~\ref{alg:line}) with $m=10$ evenly spaced points, 
corresponding to eight additional evaluations per step.

Recognizing the impact of the step length on MCMC methods, we followed \citet{du2023reduce} by parameterizing it as a function of the beta-schedule. For Langevin dynamics, the stepsize is given by $\langStep_t = a \beta_t^b$, and for HMC, the leapfrog stepsize is denoted $\varepsilon_t = a \beta_t^b$. To ensure a fair comparison across all baselines and proposed methods, we conducted a seeded parameter search over a predefined range of $(a,b)$ values shared across all variants. Parameters were selected by minimizing FID on a validation subset, with classification accuracy monitored as a secondary metric. The search revealed that unadjusted samplers (U-LA, U-HMC) achieved their best performance with comparatively smaller step sizes. In contrast, the adjusted variants (both energy and score) attained their lowest FID at larger step sizes. For these settings, the average empirical acceptance rates were approximately 5\% for HMC and 2\% for LA, consistent across both energy and score parameterizations.

The results are shown in Table~\ref{tab:exp:cifar100}. Average accuracy is obtained using a separate classifier trained exclusively on noise-free pairs $(\x_0, y)$, following the VGG-13-BN \mbox{architecture~\citep{simonyan2014very}.} The table shows a general trend of improvement over the baseline reverse process when additional MCMC steps are added. In particular, the MH-corrected samplers LA and HMC show significant improvements in FID scores, which are arguably the more important metric for image generation.

\begin{table}[H]
\caption{Accuracy and FID score for classifier-full guidance on CIFAR-100. 
The metrics are based on 50k generated samples for each sampling method with both energy and score models. 
The upward (\textuparrow) and downward (\textdownarrow) arrows indicate that higher and lower values are better, respectively. 
Bold numbers indicate the best performance in each metric for both Energy and Score samplers.}
\begin{tabularx}{\textwidth}{CCCC}
\toprule
 & \textbf{Sampler} & \textbf{Accuracy [\%]~\textuparrow}
 & \textbf{FID~\textdownarrow} \\ 
\midrule
\multirow{5}{*}{Energy} 
& Reverse   & 72.6 & 33.4 \\
& U-LA      & \boldmath{${87.3}$}
 & 24.6 \\
& LA        & 80.0 & 12.7 \\
& U-HMC     & 87.2 & 25.4 \\
& HMC       & 84.9 & \boldmath{${12.4}$} \\
\midrule
\multirow{5}{*}{Score}
& Reverse   & $74.2$ & $31.8$ \\
& U-LA      & \boldmath{${82.9}$} & $25.9$\\
& LA-10L     & $75.2$ & $15.5$ \\
& U-HMC     & $79.0$ & $28.6$ \\
& HMC-3C    & $75.8$ & \boldmath{${13.3}$} \\
\bottomrule
\end{tabularx}
\label{tab:exp:cifar100}
\end{table}

Comparing the score and energy parameterizations, their performances share similar characteristics. Interestingly, the reverse process favors the score parameterization, supporting the claim that this less restricted approach better models the score function. However, the energy parameterization sees larger improvements from the added MCMC steps. This indicates, perhaps, that direct energy estimation provides a better correction step compared to our method of approximating the pseudo-energy difference from $\epsP$. Although the energy-based method performs slightly better in this setting, our MH-corrected sampling methods achieve comparable improvements without requiring an energy model.

\textbf{ImageNet:}  
For ImageNet, training diffusion models from scratch requires substantial computational resources, so we relied on widely used pre-trained models. Score-based models are publicly available through the OpenAI GitHub repository\footnote{\url{https://github.com/openai/guided-diffusion}} as provided by \citet{dhariwal2021diffbeatgan}. However, to our knowledge, there are currently no equivalent publicly available, pre-trained energy-parameterized models at this scale. This limitation highlights the practical value of our approach: by operating directly on pre-trained score models, our method enables MCMC-based refinement even in large-scale settings where explicit energy parameterizations are not readily available. Given the high computational demands of large-scale diffusion models, we focused solely on evaluating HMC and compared it to the reverse process. The HMC sampler adds $\mcSteps=2$ MCMC steps per diffusion step $t$, with each step consisting of three leapfrog steps, following the setup of the guided diffusion experiment in~\citet{du2023reduce}. For the trapezoidal rule, we used the curve-based integration from Algorithm~\ref{alg:curve} with $m=4$.
The step length parameterization and tuning follow the same procedure as in CIFAR-100.

The results can be seen in Table~\ref{tab:exp:imagenet}. Accuracy metrics are computed using a pre-trained RegNetX-8.0GF \cite{radosavovic2020designing} classifier. The reverse process and HMC perform very similarly in average accuracy, but our method shows a slight improvement in top 5 average accuracy. HMC obtains a significantly better FID score.

\begin{table}[H]
%\centering
\caption{Average accuracy, top 5 accuracy, and FID score for classifier-full guidance on ImageNet. 
The metrics are based on 50k generated samples for both sampling methods with score parameterizations. 
The upward (\textuparrow) and downward (\textdownarrow) arrows indicate that higher and lower values are better, respectively. 
Bold numbers indicate the best performance in each metric.}
\begin{tabularx}{\textwidth}{CCCCC}
\toprule
 & \textbf{Sampler} & \textbf{Acc [\%]\textuparrow}
 & \textbf{Acc-5 [\%]~\textuparrow} & \textbf{FID~\textdownarrow} \\ 
\midrule
\multirow{2}{*}{Score}
& Reverse   & \boldmath{${50.0}$}
 & $83.9$ & $14.5$ \\
& HMC-4C    & $49.9$ & \boldmath{${85.1}$} & \boldmath{${11.6}$} \\
\bottomrule
\end{tabularx}
\label{tab:exp:imagenet}
\end{table}
\subsection{Image Tapestry}

We conduct a final, image tapestry experiment, similar to \cite{du2023reduce} and based on their code\footnote{\url{https://github.com/yilundu/reduce_reuse_recycle}}, 
with only minor modifications to incorporate our MH-like correction. The goal is to generate a coherent image composed of spatially localized content, each region conditioned on different prompts. This task involves both classifier-free guidance and model composition—specifically the combination of multiple overlapping text-to-image diffusion models, each responsible for a portion of the scene.

We use a pre-trained DeepFloyd-IF model\footnote{\url{https://huggingface.co/DeepFloyd/IF-I-XL-v1.0}}
 as the base diffusion model. To refine the generated samples, we apply Langevin dynamics with our MH-like correction. For each diffusion step ($T = 100$), we include 15 additional Langevin steps. The pseudo-energy difference is approximated via line integration using three additional evaluation points per step. We set the classifier-free guidance scale to $\lambda = 20.0$.

\begin{figure}[H]

    \begin{minipage}{0.39\textwidth}

        \includegraphics[width=\linewidth]{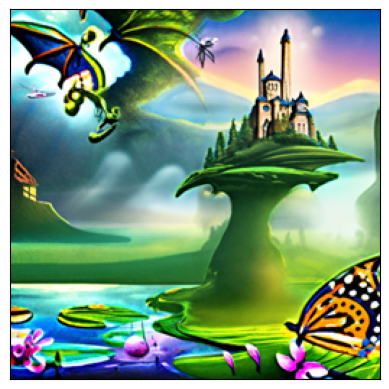}
        \vspace{0.3em}
        \centering
        (\textbf{a})
        \label{fig:tapestry}
    \end{minipage}
    \hfill
    \begin{minipage}{0.50\textwidth}

        \includegraphics[width=\linewidth]{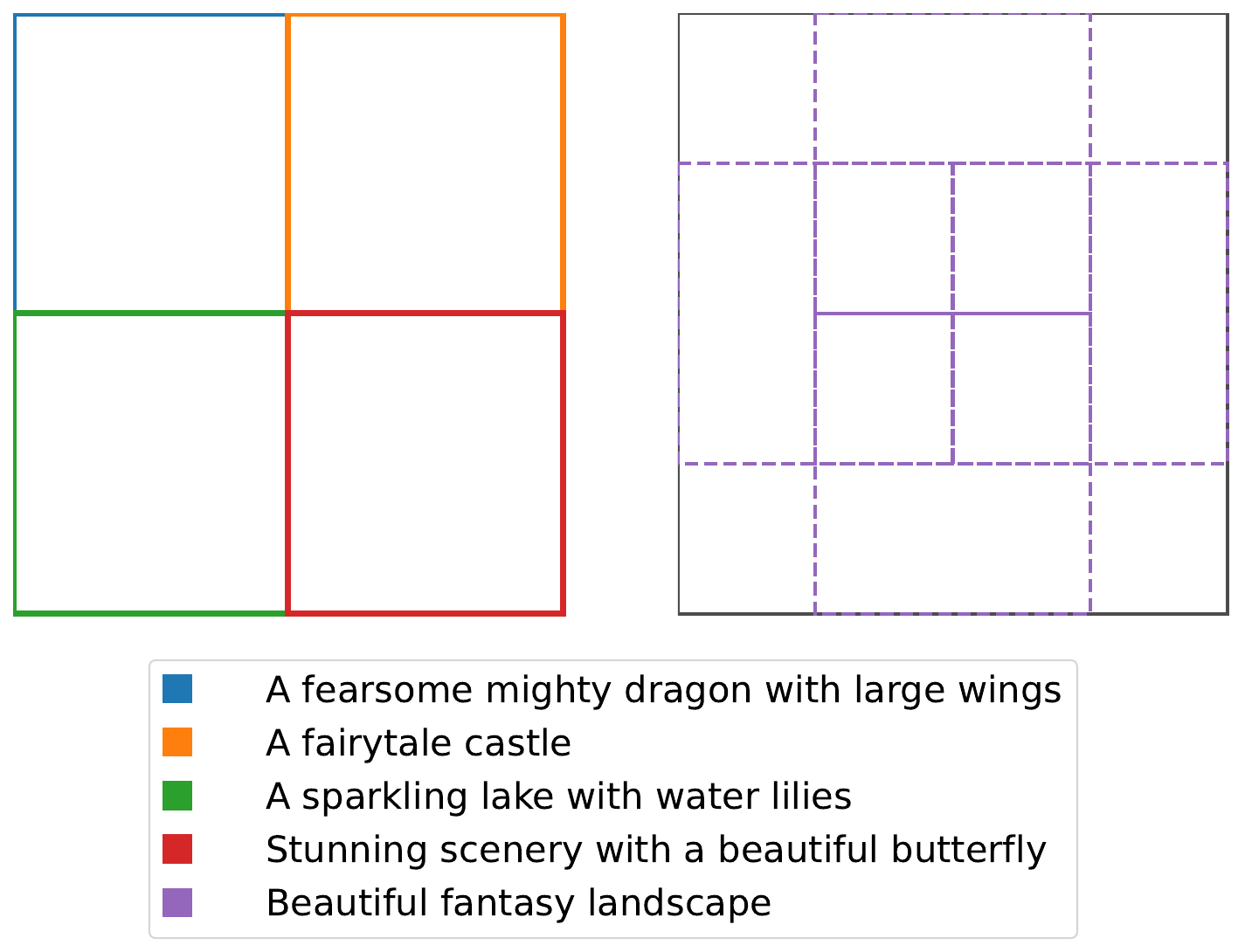}
        \vspace{0.3em}
         \centering
       (\textbf{b})
        \label{fig:tapestry_content}
    \end{minipage}

    \caption{
        In (\textbf{a}), the generated tapestry image is shown with different content at various locations.
        In (\textbf{b}), the specified content and their positions are illustrated.
    }
    \label{fig:combined}
\end{figure}

The resulting image is presented in Figure~\ref{fig:combined}a, which showcases the generated tapestry with different regions displaying distinct visual content. Figure~\ref{fig:combined}b provides a schematic overview of the used prompts and their spatial layout. In total, nine content regions are specified: four located in the corners of the image, each with unique prompts, and five overlapping in the center, all guided by the same prompt to create a unified visual theme.
\section{Discussion}

The choice between score and energy parameterizations remains an intriguing and nuanced topic within diffusion-based generative modeling. In this work, we provided additional empirical evidence suggesting that the score parameterization performs better in the standard reverse process.

At the same time, we showed that performance gains often attributed to the energy parameterization can, in fact, be recovered within a score-based framework. This is achieved by approximating pseudo-energy differences using a line integral of the model's noise predictions. Notably, this allows us to incorporate MH-like correction steps into a variety of MCMC samplers—without the need to explicitly train an energy-based model—{achieving relative improvements in sample quality of a similar magnitude. This represents a crucial practical benefit, as score-based models are fundamentally easier to train and scale to high-dimensional datasets compared to energy-based parameterizations.}

A particularly interesting observation is that using a curve composed only of model evaluations from the HMC sampler appears to perform on par with using a straight-line path. This suggests that the proposed correction comes at virtually no additional computational cost in this case. However, it is worth noting that in higher-dimensional settings, additional intermediate points along the integration path may be required to maintain accuracy, which could increase the computational burden. This challenge might be addressed through more efficient numerical integration techniques, or by working in a lower-dimensional latent space, as in latent diffusion models. One persistent drawback of the energy parameterization is that it always requires an explicit gradient computation to recover the score function.

Another important consideration is computational cost. As we observed in the experiments, adding MCMC updates on top of the reverse diffusion process introduces a clear overhead in terms of additional score evaluations. Our approach is therefore not intended as a replacement for accelerated solvers, e.g., PF-ODE, which aim to minimize the number of function evaluations. Instead, we view MH-like corrections as an orthogonal contribution: they can be applied on top of any diffusion model and any sampler. In this sense, our method is complementary to existing work, targeting improved sample quality while remaining agnostic to the choice of backbone or solver.

{Our empirical step length parameter search on CIFAR-100 highlights a qualitative difference between adjusted and unadjusted samplers. The unadjusted variants (U-LA, U-HMC), which lack a rejection mechanism, achieved their best performance only within a relatively narrow range of smaller step sizes. Outside this regime, image quality deteriorated rapidly during sampling. In contrast, the adjusted variants consistently produced semantically coherent samples across a broader portion of the explored parameter range. Empirically, the rejection step appears to mitigate the destabilizing effect of overly aggressive proposals, effectively filtering out unfavorable transitions. For the best performing configurations, the adjusted samplers exhibited relatively low average acceptance rates (around 2 or 5\%). Notably, both energy-based and score-based adjusted samplers achieved peak performance at similar acceptance levels, indicating comparable dynamical behavior between explicit and pseudo-energy formulations.}

In line with prior work such as~\citet{du2023reduce}, our experimental design deliberately focused on widely used baselines rather than the strongest available backbones. This choice allowed us to isolate the effect of the MH-like correction without confounding factors from architectural or solver improvements. While this comes at the cost of absolute FID values that are below the state of the art, the relative improvements we observe consistently demonstrate the benefit of our approach. We expect that applying our method to stronger architectures will yield proportionally similar gains, but leave this as an exciting direction for future work.

One limitation of the score parameterization is that the learned vector field is not guaranteed to be conservative, and therefore our line-integral construction does not, in general, yield a true Metropolis--Hastings correction. {Consequently, the resulting acceptance rule should not be interpreted as guaranteeing detailed balance with respect to the target distribution. In practice, the samples are generated by the sequential application of the reverse diffusion updates together with MCMC steps and the practically motivated MH-like acceptance rule. The resulting distribution should therefore be understood as the implicit distribution induced by this procedure.} Importantly, our construction recovers a true Metropolis--Hastings correction in the special case where the score is conservative, thereby aligning with the energy-based formulation. More broadly, recent work by Horvat and Pfister~\citep{horvat2024gauge} highlights that score representations may admit non-conservative components without necessarily undermining their effectiveness for generative modeling. This perspective aligns with the empirical success of score-based generative models that operate without explicitly modeling an energy function. Consistent with this perspective, our MH-like correction mechanism—though built on a generally non-conservative field—yields consistent improvements across the evaluated datasets and samplers when applied to the reverse process.

\textls[-15]{To further assess the practical impact of the non-conservative nature of the learned score field, we conducted a series of empirical sanity checks (detailed in \mbox{Appendices~\ref{sec:app:mnist_line_curve} and \ref{sec:appcifar100}).} First, to examine potential path dependence, we compared pseudo-energy differences obtained from straight-line and curved integration paths. On both  trained MNIST models and the higher-dimensional CIFAR-100 model, the observed discrepancies were small relative to the magnitude of the pseudo-energy differences. This indicates limited path sensitivity at the proposal scale relevant for our MCMC refinement steps. Furthermore, we evaluated the numerical stability of the line-integral approximation via a mesh convergence analysis on CIFAR-100 (detailed in Appendix \ref{sec:app_mesh}). We observed rapid convergence of the discretization scheme: the median symmetric relative discrepancy falls below 1\% when using $n \ge 10$ integration points compared to a higher-resolution reference. Together, these empirical results suggest that the learned score behaves approximately conservatively on the local scales relevant for MCMC proposals, as small variations in the integration path yield similar pseudo-energy differences.}

Still, the lack of theoretical guarantees may explain the slightly superior performance of the energy parameterization observed in the CIFAR-100 experiment. Developing better techniques for estimating pseudo-energy differences from score-based models—without requiring an explicitly trained energy function—thus remains a highly relevant and promising direction for future research.
\section{Conclusions}

We introduced a method for extending the reverse diffusion process with MCMC sampling based on an MH-like correction step computed from the score function. This approach enables improved sampling for composed diffusion models without requiring an energy-based parameterization.

While previous work,~\citet{du2023reduce}, demonstrated the benefits of MH correction under an energy parameterization, our method instead defines a pseudo-energy difference derived from the score, estimated via numerical integration. This allows us to apply MH-like corrections in the score-based setting—by far the most common in practice—and thereby make use of existing pre-trained diffusion models for composition tasks.

Our method can reuse intermediate evaluations from samplers such as HMC to compute the correction with little to no additional cost. In general, the accuracy of the MH-like correction depends on the numerical integration of the score, which may require more intermediate points as the dimensionality increases. While this can introduce some overhead, energy-based methods incur their own costs, such as differentiating the energy function. In practice, our corrected score-based samplers consistently {yield relative improvements of a similar magnitude to} energy-based methods across a range of tasks, making them a practical alternative in settings where score-based models are already available.

Overall, our work extends the applicability of corrected MCMC sampling to the broad class of score-based diffusion models and opens the door to more flexible and modular composition of generative models.
\section*{Funding}
This research was funded by the Wallenberg AI, Autonomous Systems and Software Program (WASP), funded by the Knut and Alice Wallenberg Foundation. The computations were enabled by resources provided by the National Academic Infrastructure for Supercomputing in Sweden (NAISS), partially funded by the Swedish Research Council (grant no. 2022-06725).

\section*{Data Availability}
All datasets used in this study are either publicly available from the sources cited in the manuscript, or can be fully reproduced from the code provided.

\section*{Acknowledgments}
During the preparation of this manuscript, the authors used ChatGPT for assistance in improving the clarity and readability of the manuscript. The authors reviewed and edited all AI-assisted content and take full responsibility for the final version of the manuscript.

\section*{Conflicts of Interest}
The authors declare no conflicts of interest.

\bibliography{references}

@article{ho2020denoising,
  title={Denoising diffusion probabilistic models},
  author={Ho, Jonathan and Jain, Ajay and Abbeel, Pieter},
  journal={Advances in neural information processing systems},
  volume={33},
  pages={6840--6851},
  year={2020}
}

@inproceedings{sohl2015deep,
  title={Deep unsupervised learning using nonequilibrium thermodynamics},
  author={Sohl-Dickstein, Jascha and Weiss, Eric and Maheswaranathan, Niru and Ganguli, Surya},
  booktitle={International conference on machine learning},
  pages={2256--2265},
  year={2015},
  organization={PMLR}
}

@article{song2019generative,
  title={Generative modeling by estimating gradients of the data distribution},
  author={Song, Yang and Ermon, Stefano},
  journal={Advances in neural information processing systems},
  volume={32},
  year={2019}
}

@inproceedings{
song2021scorebased,
title={Score-Based Generative Modeling through Stochastic Differential Equations},
author={Yang Song and Jascha Sohl-Dickstein and Diederik P Kingma and Abhishek Kumar and Stefano Ermon and Ben Poole},
booktitle={International Conference on Learning Representations},
year={2021},
}

@inproceedings{
salimans2021should,
title={Should {EBM}s model the energy or the score?},
author={Tim Salimans and Jonathan Ho},
booktitle={Energy Based Models Workshop - ICLR 2021},
year={2021},
url={https://openreview.net/forum?id=9AS-TF2jRNb}
}

@inproceedings{
ho2021classifierfree,
title={Classifier-Free Diffusion Guidance},
author={Jonathan Ho and Tim Salimans},
booktitle={NeurIPS 2021 Workshop on Deep Generative Models and Downstream Applications},
year={2021},
}

@inproceedings{
chung2023reconstrguidance,
title={Diffusion Posterior Sampling for General Noisy Inverse Problems},
author={Hyungjin Chung and Jeongsol Kim and Michael Thompson Mccann and Marc Louis Klasky and Jong Chul Ye},
booktitle={The Eleventh International Conference on Learning Representations },
year={2023},
}

@inproceedings{ho2022guidance,
 author = {Ho, Jonathan and Salimans, Tim and Gritsenko, Alexey and Chan, William and Norouzi, Mohammad and Fleet, David J},
 booktitle = {Advances in Neural Information Processing Systems},
 editor = {S. Koyejo and S. Mohamed and A. Agarwal and D. Belgrave and K. Cho and A. Oh},
 pages = {8633--8646},
 publisher = {Curran Associates, Inc.},
 title = {Video Diffusion Models},
 volume = {35},
 year = {2022}
}

@article{dhariwal2021diffbeatgan,
  title={Diffusion models beat {GAN}s on image synthesis},
  author={Dhariwal, Prafulla and Nichol, Alexander},
  journal={Advances in neural information processing systems},
  volume={34},
  pages={8780--8794},
  year={2021}
}

@article{li2022competition,
  title={Competition-level code generation with alphacode},
  author={Li, Yujia and Choi, David and Chung, Junyoung and Kushman, Nate and Schrittwieser, Julian and Leblond, R{\'e}mi and Eccles, Tom and Keeling, James and Gimeno, Felix and Dal Lago, Agustin and others},
  journal={Science},
  volume={378},
  number={6624},
  pages={1092--1097},
  year={2022},
  publisher={American Association for the Advancement of Science}
}

@article{gungor2023adaptive,
  title={Adaptive diffusion priors for accelerated MRI reconstruction},
  author={G{\"u}ng{\"o}r, Alper and Dar, Salman UH and {\"O}zt{\"u}rk, {\c{S}}aban and Korkmaz, Yilmaz and Bedel, Hasan A and Elmas, Gokberk and Ozbey, Muzaffer and {\c{C}}ukur, Tolga},
  journal={Medical Image Analysis},
  pages={102872},
  year={2023},
  publisher={Elsevier}
}

@inproceedings{wynn2023diffusionerf,
  title={Diffusio{N}e{RF}: Regularizing neural radiance fields with denoising diffusion models},
  author={Wynn, Jamie and Turmukhambetov, Daniyar},
  booktitle={Proceedings of the IEEE/CVF Conference on Computer Vision and Pattern Recognition},
  pages={4180--4189},
  year={2023}
}

@article{saharia2022photorealistic,
  title={Photorealistic text-to-image diffusion models with deep language understanding},
  author={Saharia, Chitwan and Chan, William and Saxena, Saurabh and Li, Lala and Whang, Jay and Denton, Emily L and Ghasemipour, Kamyar and Gontijo Lopes, Raphael and Karagol Ayan, Burcu and Salimans, Tim and others},
  journal={Advances in Neural Information Processing Systems},
  volume={35},
  pages={36479--36494},
  year={2022}
}

@article{jacobs1991adaptive,
  title={Adaptive mixtures of local experts},
  author={Jacobs, Robert A and Jordan, Michael I and Nowlan, Steven J and Hinton, Geoffrey E},
  journal={Neural computation},
  volume={3},
  number={1},
  pages={79--87},
  year={1991},
  publisher={MIT Press}
}

@article{hinton2002training,
  title={Training products of experts by minimizing contrastive divergence},
  author={Hinton, Geoffrey E},
  journal={Neural computation},
  volume={14},
  number={8},
  pages={1771--1800},
  year={2002},
  publisher={MIT Press}
}

@article{mayraz2000recognizing,
  title={Recognizing hand-written digits using hierarchical products of experts},
  author={Mayraz, Guy and Hinton, Geoffrey E},
  journal={Advances in neural information processing systems},
  volume={13},
  year={2000}
}

@InProceedings{du2023reduce,
  title = 	 {Reduce, {R}euse, {R}ecycle: Compositional Generation with Energy-Based Diffusion Models and {MCMC}},
  author =       {Du, Yilun and Durkan, Conor and Strudel, Robin and Tenenbaum, Joshua B. and Dieleman, Sander and Fergus, Rob and Sohl-Dickstein, Jascha and Doucet, Arnaud and Grathwohl, Will Sussman},
  booktitle = 	 {Proceedings of the 40th International Conference on Machine Learning},
  pages = 	 {8489--8510},
  year = 	 {2023},
  editor = 	 {Krause, Andreas and Brunskill, Emma and Cho, Kyunghyun and Engelhardt, Barbara and Sabato, Sivan and Scarlett, Jonathan},
  volume = 	 {202},
  series = 	 {Proceedings of Machine Learning Research},
  month = 	 {23--29 Jul},
  publisher =    {PMLR},
}

@inproceedings{aghajanyan2023scaling,
author = {Aghajanyan, Armen and Yu, Lili and Conneau, Alexis and Hsu, Wei-Ning and Hambardzumyan, Karen and Zhang, Susan and Roller, Stephen and Goyal, Naman and Levy, Omer and Zettlemoyer, Luke},
title = {Scaling laws for generative mixed-modal language models},
year = {2023},
publisher = {JMLR},
booktitle = {Proceedings of the 40th International Conference on Machine Learning},
articleno = {13},
numpages = {15},
series = {ICML'23}
}

@article{metropolis1953mh,
author = {Metropolis, Nicholas and Rosenbluth, Arianna W. and Rosenbluth, Marshall N. and Teller, Augusta H.},
journal = {The Journal of Chemical Physics},
number = {6},
pages = {1087--1092},
title = {Equation of State Calculations by Fast Computing Machines},
volume = {21},
year = {1953}
}

@article{hasting1970mh,
author = {Hastings, W. K.},
journal = {Biometrika},
number = {1},
pages = {97--109},
title = {{M}onte {C}arlo Sampling Methods Using {M}arkov Chains and Their Applications},
volume = {57},
year = {1970}
}

@book{
neal1996bayes,
Author = {Neal, Radford M. and Diggle, P. and Fienberg, S.},
ISBN = {9781461207450},
Publisher = {Springer New York},
Series = {Lecture Notes in Statistics Ser.: v.118},
Title = {Bayesian Learning for Neural Networks.},
Year = {1996},
}

@article{roberts2002mala,
Author = {Roberts, G. O. and Stramer, O.},
ISSN = {13875841},
Journal = {Methodology \& Computing in Applied Probability},
Number = {4},
Pages = {337 - 357},
Title = {Langevin Diffusions and {M}etropolis--{H}astings Algorithms.},
Volume = {4},
Year = {2002},
}

@article{duane1987hmc,
title = {Hybrid {M}onte {C}arlo},
journal = {Physics Letters B},
volume = {195},
number = {2},
pages = {216-222},
year = {1987},
issn = {0370-2693},
author = {Simon Duane and A.D. Kennedy and Brian J. Pendleton and Duncan Roweth},
}

@inproceedings{nichol2021improved,
  title={Improved denoising diffusion probabilistic models},
  author={Nichol, Alexander Quinn and Dhariwal, Prafulla},
  booktitle={International Conference on Machine Learning},
  pages={8162--8171},
  year={2021},
  organization={PMLR}
}

@inproceedings{liu2022compositional,
  title={Compositional visual generation with composable diffusion models},
  author={Liu, Nan and Li, Shuang and Du, Yilun and Torralba, Antonio and Tenenbaum, Joshua B},
  booktitle={European Conference on Computer Vision},
  pages={423--439},
  year={2022},
  organization={Springer}
}

@article{heusel2017gans,
  title={Gans trained by a two time-scale update rule converge to a local nash equilibrium},
  author={Heusel, Martin and Ramsauer, Hubert and Unterthiner, Thomas and Nessler, Bernhard and Hochreiter, Sepp},
  journal={Advances in neural information processing systems},
  volume={30},
  year={2017}
}

@article{brown2020language,
  title={Language models are few-shot learners},
  author={Brown, Tom and Mann, Benjamin and Ryder, Nick and Subbiah, Melanie and Kaplan, Jared D and Dhariwal, Prafulla and Neelakantan, Arvind and Shyam, Pranav and Sastry, Girish and Askell, Amanda and others},
  journal={Advances in neural information processing systems},
  volume={33},
  pages={1877--1901},
  year={2020}
}

@inproceedings{brock2018large,
title={Large Scale {GAN} Training for High Fidelity Natural Image Synthesis},
author={Andrew Brock and Jeff Donahue and Karen Simonyan},
booktitle={International Conference on Learning Representations},
year={2019},
}

@inproceedings{ludke2023diff_tpp,
	title = {Add and Thin: Diffusion for Temporal Point Processes},
	shorttitle = {Add and Thin},
	booktitle = {Thirty-seventh Conference on Neural Information Processing Systems},
	author = {L{\"u}dke, David and Bilo{\v s}, Marin and Shchur, Oleksandr and Lienen, Marten and Günnemann, Stephan},
    year = {2023},
}

@misc{wang2024neural,
      title={Neural Network Diffusion}, 
      author={Kai Wang and Zhaopan Xu and Yukun Zhou and Zelin Zang and Trevor Darrell and Zhuang Liu and Yang You},
      year={2024},
      eprint={2402.13144},
      archivePrefix={arXiv},
      primaryClass={cs.LG}
}

@Techreport{krizhevsky2009cifar,
 author = {Krizhevsky, Alex and Hinton, Geoffrey},
 address = {Toronto, Ontario},
 institution = {University of Toronto},
 number = {0},
 publisher = {Technical report, University of Toronto},
 title = {Learning multiple layers of features from tiny images},
 year = {2009},
 title_with_no_special_chars = {Learning multiple layers of features from tiny images},
 url = {https://www.cs.toronto.edu/~kriz/learning-features-2009-TR.pdf}
}

@inproceedings{deng2009imagenet,
  title={Imagenet: A large-scale hierarchical image database},
  author={Deng, Jia and Dong, Wei and Socher, Richard and Li, Li-Jia and Li, Kai and Fei-Fei, Li},

  booktitle={2009 IEEE conference on computer vision and pattern recognition},
  pages={248--255},
  year={2009},
  organization={IEEE}
}

@article{simonyan2014very,
  title={Very deep convolutional networks for large-scale image recognition},
  author={Simonyan, Karen and Zisserman, Andrew},
  journal={arXiv preprint arXiv:1409.1556},
  year={2014}
}

@inproceedings{radosavovic2020designing,
  title={Designing network design spaces},
  author={Radosavovic, Ilija and Kosaraju, Raj Prateek and Girshick, Ross and He, Kaiming and Doll{\'a}r, Piotr},
  booktitle={Proceedings of the IEEE/CVF conference on computer vision and pattern recognition},
  pages={10428--10436},
  year={2020}
}

@article{lecun2006tutorial,
  title={A tutorial on energy-based learning},
  author={LeCun, Yann and Chopra, Sumit and Hadsell, Raia and Ranzato, M and Huang, Fujie and others},
  journal={Predicting structured data},
  volume={1},
  number={0},
  year={2006}
}

@article{horvat2024gauge,
  title={On gauge freedom, conservativity and intrinsic dimensionality estimation in diffusion models},
  author={Horvat, Christian and Pfister, Jean-Pascal},
  journal={arXiv preprint arXiv:2402.03845},
  year={2024}
}

@article{mnist,
  author={Deng, Li},
  journal={IEEE Signal Processing Magazine}, 
  title={The MNIST Database of Handwritten Digit Images for Machine Learning Research [Best of the Web]}, 
  year={2012},
  volume={29},
  number={6},
  pages={141-142},
  doi={10.1109/MSP.2012.2211477}}

@article{neal2001annealed,
  title={Annealed importance sampling},
  author={Neal, Radford M},
  journal={Statistics and computing},
  volume={11},
  pages={125--139},
  year={2001},
  publisher={Springer}
}

@misc{jax2018github,
  author       = {Bradbury, James and Frostig, Roy and Hawkins, Peter and
                  Johnson, Matthew James and Leary, Chris and Maclaurin, Dougal and
                  Necula, George and Paszke, Adam and VanderPlas, Jake and
                  Wanderman-Milne, Skye and Zhang, Qiao},
  title        = {{JAX}: composable transformations of {P}ython+{N}um{P}y programs},
  howpublished = {\url{https://github.com/google/jax}},
  year         = {2018},
  note         = {Version 0.4.30}
}

@article{geffner2021mcmc,
  title={MCMC variational inference via uncorrected Hamiltonian annealing},
  author={Geffner, Tomas and Domke, Justin},
  journal={Advances in Neural Information Processing Systems},
  volume={34},
  pages={639--651},
  year={2021}
}

@article{besag1994mala,
  author  = {Besag, Julian},
  title   = {Comments on ``{R}epresentations of knowledge in complex systems'' by U. Grenander and M. I. Miller},
  journal = {Journal of the Royal Statistical Society: Series B (Methodological)},
  volume  = {56},
  pages   = {591--592},
  year    = {1994}
}

@article{roberts1996exponential,
  title   = {Exponential convergence of Langevin distributions and their discrete approximations},
  author  = {Roberts, Gareth O. and Tweedie, Richard L.},
  journal = {Bernoulli},
  volume  = {2},
  number  = {4},
  pages   = {341--363},
  year    = {1996},
  month   = dec
}
\bibliographystyle{plainnat}

\newpage
\newpage
\onecolumn
\appendix

\section{Experimental Details}
\label{sec:expdetails}

Here we provide more details about our different conducted experiments: Evaluating pseudo-energy differences, two-dimensional composition, guided diffusion, and \mbox{image tapestry.}

The earlier experiments were conducted on a machine equipped with an {NVIDIA GeForce RTX 3060}, while the later experiments were run on a computing cluster with {NVIDIA A100 Tensor Core GPUs}.

\subsection{Evaluating Pseudo-Energy Difference} \label{sec:app:ped}

All models in this section were trained with the Adam optimizer using a learning rate of $10^{-3}$, together with a StepLR scheduler with step size $1$ 
and decay factor $0.99$. 

\textbf{2D Gaussian:}
We generated samples from a bivariate Gaussian distribution with mean $\boldsymbol{\mu} = (2, 0)^\top$ and covariance $\boldsymbol{\Sigma} = 0.1 I$, where $I$ is the identity matrix. 

The diffusion models use $T = 100$ timesteps, with the noise schedule $\beta_t$ following the cosine schedule proposed in \cite{nichol2021improved}.

We used the same neural network architectures as the base for both the score and energy models. It is a residual network consisting of a linear layer (dim $2 \rightarrow 128$) followed by four blocks, and concluding with a linear layer (dim $128 \rightarrow 2$).
Within each block, the input $x$ passes through a normalization layer, a SiLU activation, and a linear layer (dim $128 \rightarrow 256$). Subsequently, it is sumpplemenyed with an embedded $t$ (dim 32) that has undergone a linear layer transformation (dim $32 \rightarrow 256$). The resulting sum passes through a SiLU activation and is further processed by a linear layer (dim $256 \rightarrow 256$). After that, another SiLU activation is applied, followed by a final linear layer (dim $256 \rightarrow 128$). The output of this linear layer is then added to the original input $x$ within the block. The embedding of $t$ is also learnable. 

\textbf{MNIST:}
The diffusion models use $T = 1000$ timesteps, with the noise schedule $\beta_t$ following the cosine schedule. 
For score parameterization, we trained a UNet-based architecture adapted to $28 \times 28$ grayscale images. 
The network uses a time embedding of dimension $112$, implemented as sinusoidal position embeddings followed by two fully connected layers with GELU activations. 
The model begins with a $1 \times 1$ convolution mapping the input image to dimension $28$ and proceeds through three down-sampling stages, a middle block, and three up-sampling stages. 
Each down/upsampling stage consists of two residual blocks with time conditioning, a linear or full attention layer, and either a strided convolution (downsampling) or nearest-neighbor upsampling with convolution (upsampling). 
The middle block contains two ResNet blocks and one full attention layer. Skip connections are applied between corresponding encoder and decoder layers, following the standard UNet design. 
The output stage concatenates the upsampled features with the initial projection and applies a residual block followed by a $1 \times 1$ convolution to map back to the image space. 

For energy parameterization, the architecture is identical, but the output head is replaced with an energy function whose gradient defines the score. 

\subsection{Two-Dimensional Composition} %Attention: Subtitle edit.
\label{sec:expdetails:2d_comp}

The composed distribution is defined by a product of two components, a Gaussian mixture and a uniform distribution with non-zero values on
\begin{align} \label{eq:bar}
    \square = \{ x \in \mathbb{R}^2: -s_i \leq x_i \leq s_i, i = 1,2 \},
\end{align}
where $s_1$ and $s_2$ are equal to $0.2$ and $1.0$, respectively. The eight modes of the Gaussian mixture are evenly distributed on a circle with a radius of 0.5 at the angles $\frac{\pi}{4}i$ for $i = 0, \ldots, 7$, respectively. The covariance matrix at each mode is $0.03^2 \cdot I$, where $I$ is the identity matrix.

We use the same network architecture setup for score and energy as in the 2D Gaussian case (see Appendix \ref{sec:app:ped}).

% log.-likelihood or log-likelihood?
The metric log-likelihood is ill-defined as we may generate samples where the true distribution has no support (due to the uniform distribution). We address this problem by expanding the definition set of the uniform distribution and redistributing one percent of the probability mass into this extended region. The whole set is defined as \eqref{eq:bar} except $s_1=s_2=1.1$. Note that the 99 percent probability mass remains inside the original definition set $\square$.

The parameter $\beta_t$ follows the cosine schedule. For (U-)HMC, the damping coefficient is set to $0.5$, the mass diagonal matrix has all diagonal elements equal to $1$, and the stepsize for each $t$ is $0.03$. For (U-)LA, the stepsize for each $t$ is set to $0.001$.

\subsection{Guided Diffusion for CIFAR-100}
\label{sec:expdetails:cifar100_guid}

 The parameter $\beta_t$ has a linear schedule as originally proposed in \cite{ho2020denoising}. For (U)-HMC, the damping coefficient is equal to 0.9 and the diagonal elements in the mass matrix are equal to $\beta_t$ for each $t$. The values of the stepsize parameters $a$ and $b$ were determined through a simple parameter search for the different MCMC methods and they can be found in Table \ref{tab:mcmc_stepsize_cifar100}. This was performed for both the score and energy parameterizations, where the stepsize is defined as $\langStep_t = a \beta_t^b$. 
\begin{table}[H]

     \caption{The values of the stepsize parameters $a$ and $b$ obtained from a random parameter search for the different MCMC methods for both score and energy parameterization in the CIFAR-100 experiment, where the stepsize is defined as $\langStep_t = a \beta_t^b$.}
    \label{tab:mcmc_stepsize_cifar100}
    \begin{tabularx}{\textwidth}{CCCC}
\toprule
        \multirow{2}{*}{} & \multirow{2}{*}{\textbf{MCMC\vspace{-5pt}}} & \multicolumn{2}{c}{\textbf{Stepsize Parameters}} \\ \cmidrule{3-4}
         &  & \textbf{$a$} 
 & \textbf{$b$} \\ \toprule
        \multirow{4}{*}{Energy\vspace{-12pt}} & U-LA & 9.22 & 1.40 \\ 
        \cmidrule{2-4}
         & LA & 9.84 & 0.83 \\ 
         \cmidrule{2-4}
         & U-HMC & 0.26 & 1.53 \\ 
         \cmidrule{2-4}
         & HMC & 9.33 & 1.48 \\ 
         \midrule
        \multirow{4}{*}{Score\vspace{-12pt}} & U-LA & 1.96 & 1.04 \\ 
        \cmidrule{2-4}
         & LA & 9.84 & 0.83 \\
          \cmidrule{2-4}
         & U-HMC & 0.26 & 1.53 \\
          \cmidrule{2-4}
         & HMC & 4.03 & 1.34 \\ 
        \bottomrule
    \end{tabularx}
\end{table}

To complement the quantitative results in the main text, 
we report in Table~\ref{tab:nfe_cifar100} the theoretical number of function evaluations (NFE) 
per generated sample for CIFAR-100. 
We separate forward passes (FPs) and backward passes (BPs), since energy parameterizations require BPs for score evaluations whereas 
score parameterizations only require FPs. 
Counts are computed per algorithm step and correspond to the maximum evaluations 
assuming all proposals are accepted in MH.
\begin{table}[H]
 %   \centering
    \caption{Theoretical number of function evaluations (NFE) per generated sample on CIFAR-100. We report forward passes (FPs) and backward passes (BPs) separately. Counts are computed per algorithm step, independent of accept/reject outcomes.}
    \label{tab:nfe_cifar100}
    \begin{tabularx}{\textwidth}{CCCC}
\toprule
         & \textbf{Sampler} & \textbf{NFE (FPs)} & \textbf{NFE (BPs)} \\
  \midrule
        \multirow{5}{*}{Energy}
        & Reverse & 0 & 1000 \\
        & U-LA    & 0 & 6994 \\
        & LA      & 6993 & 7993 \\
        & U-HMC   & 0 & 6994 \\
        & HMC     & 2997 & 7993 \\
\midrule
        \multirow{5}{*}{Score}
        & Reverse & 1000 & 0 \\
        & U-LA    & 6994 & 0 \\
        & LA-10L      & 55,945 & 0 \\
        & U-HMC   & 6994 & 0 \\
        & HMC-3C     & 13,982 & 0 \\
    \bottomrule
    \end{tabularx}
\end{table}

\begin{figure}[H]
    %\centering
    \begin{minipage}[t]{0.48\textwidth}
        \centering
        \includegraphics[width=\textwidth]{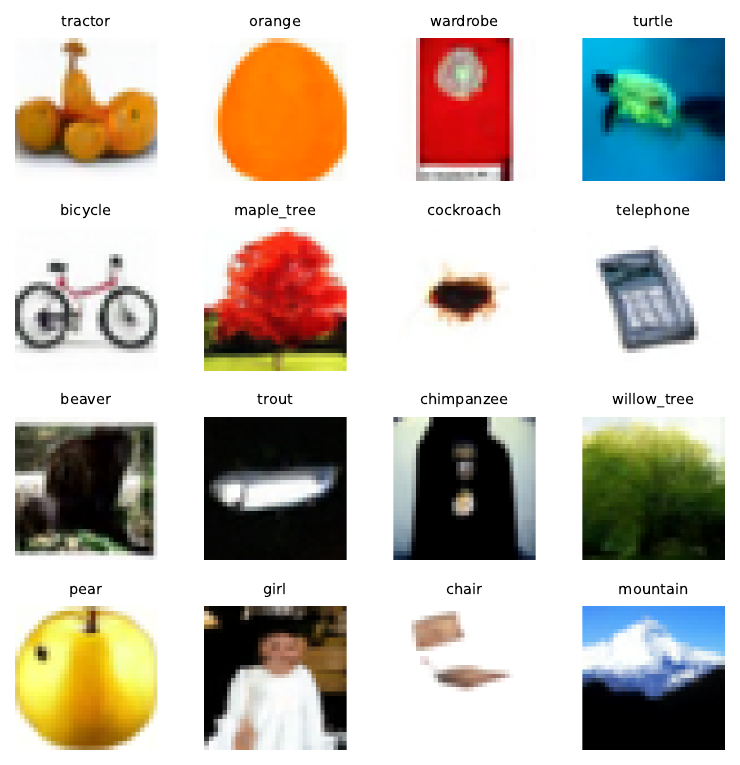}
        \vspace{0.5em}
        {\small (\textbf{a}) CIFAR-100 Reverse sampler}
        \label{fig:cifar100_reverse}
    \end{minipage}
    \hfill
    \begin{minipage}[t]{0.48\textwidth}
        \centering
        \includegraphics[width=\textwidth]{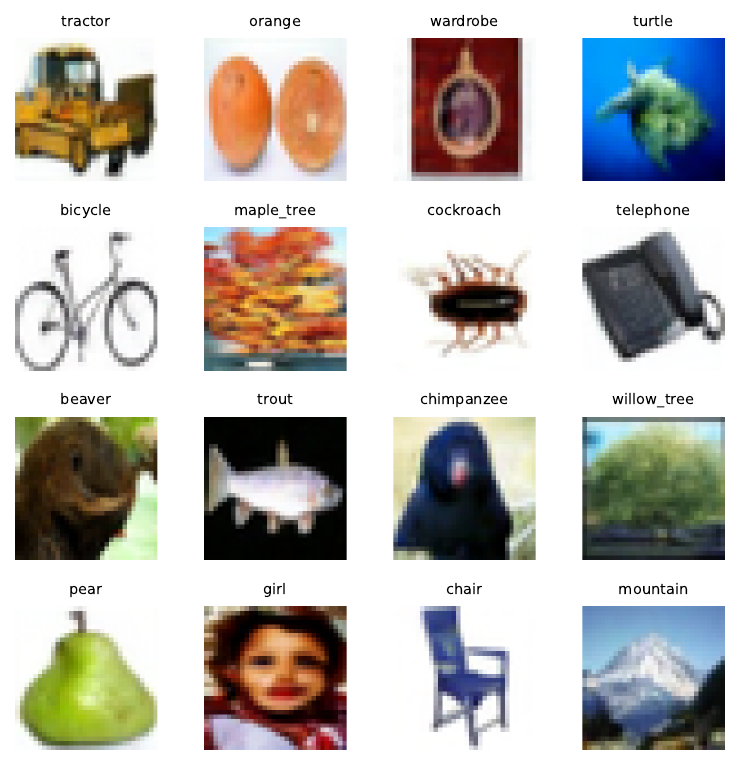}
        \vspace{0.5em}     
        {\small (\textbf{b}) CIFAR-100 HMC sampler}
        \label{fig:cifar100_hmc}
    \end{minipage}

    \caption{Generated
 CIFAR-100 samples using (\textbf{a}) reverse and (\textbf{b}) HMC.}
    \label{fig:cifar100_samples}
\end{figure}

To provide a qualitative comparison, we show representative generated samples in 
Figure~\ref{fig:cifar100_samples}. Each pair of images (a) and (b) is conditioned on 
the same class label and initialized from the same noise realization $x_T$, but generated 
with different samplers.
\vspace{-6pt}

\subsection{Guided Diffusion for ImageNet}
\label{sec:expdetails:imagenet_guid}

Again, the parameter $\beta_t$ follows a linear schedule. The hyperparameters for the HMC include a damping coefficient set to $0.9$, with the diagonal elements of the mass matrix being equal to $\beta_t$ for each $t$. The stepsize parameters for HMC, obtained from a simple parameter search, are $a=1.87$ and $b=1.51$.

The ImageNet dataset (\url{https://image-net.org/}) used to compute the FID score is available for free to researchers for non-commercial use.

For the ImageNet experiment with classifier guidance, 
only score parameterizations were considered. 
Table~\ref{tab:nfe_imagenet} reports the corresponding NFEs per generated sample, 
again separating forward (FPs) and backward (BPs) passes. 
Here we compare the baseline reverse sampler with our MH-like corrected HMC variant (HMC-6C).

\begin{figure}[H]
\centering
    \begin{minipage}{0.7\textwidth}
        \centering
        \includegraphics[width=0.85\textwidth]{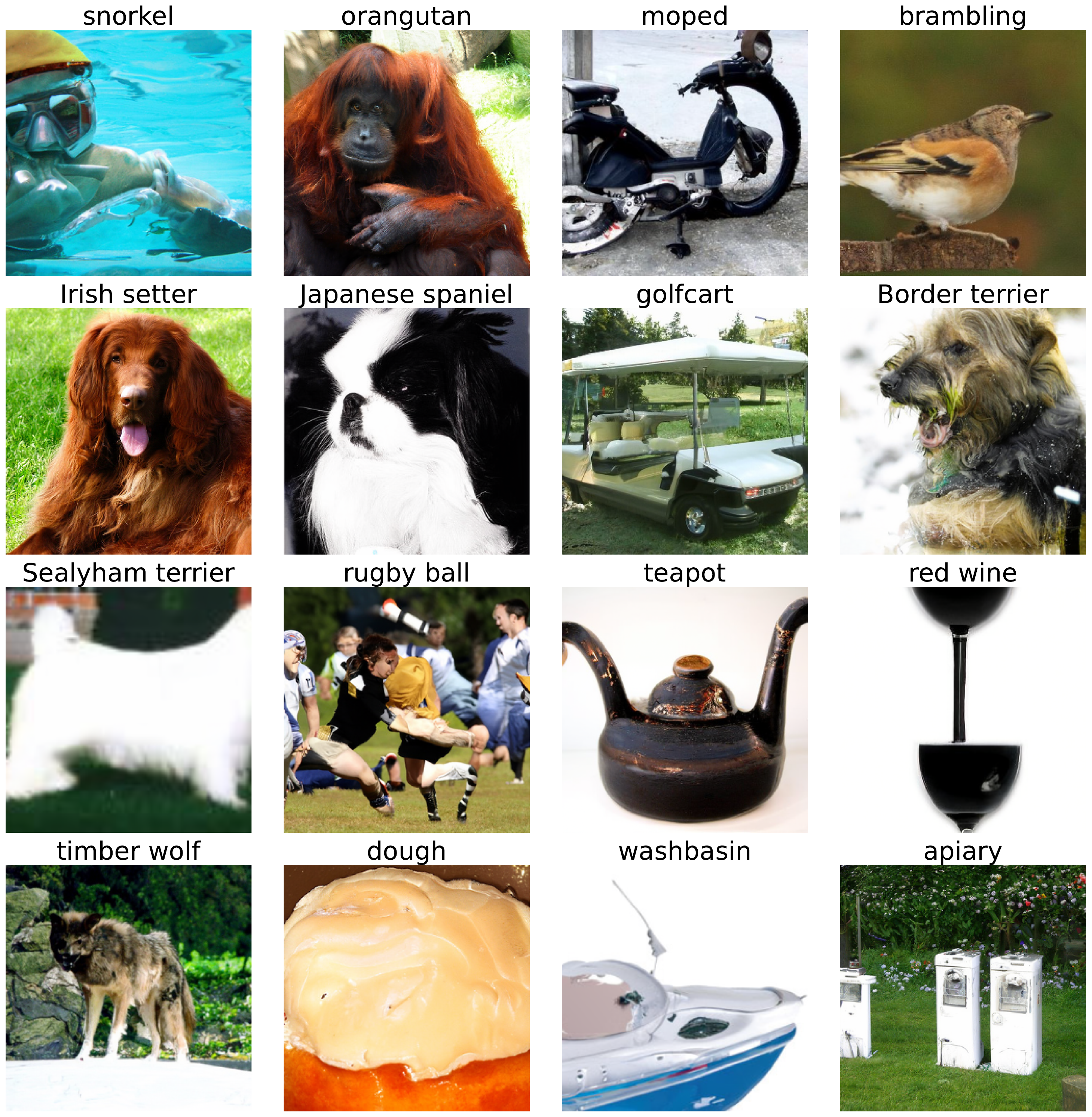}
        
        {\small (\textbf{a}) ImageNet Reverse sampler}

        \includegraphics[width=0.85\textwidth]{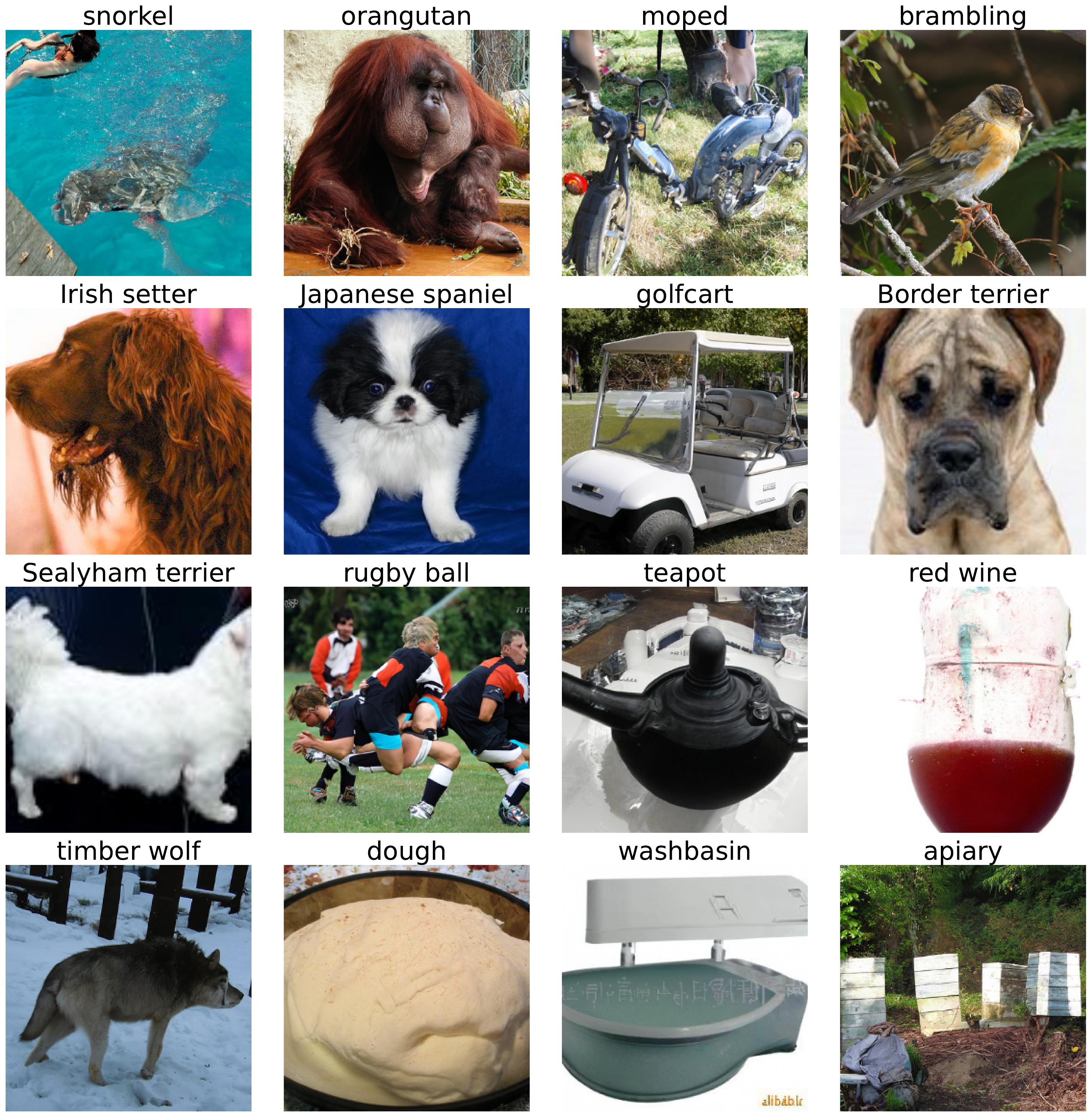}
 
        {\small (\textbf{b}) ImageNet HMC sampler}
        \label{fig:imagenet_hmc}
    \end{minipage}

    \caption{Generated ImageNet samples using (\textbf{a}) reverse and (\textbf{b}) HMC. All labels correspond to standard ImageNet class names.}
    \label{fig:imagenet_samples}
\end{figure}

For ImageNet, representative samples are shown in Figure~\ref{fig:imagenet_samples}. 
As in the CIFAR-100 case, the pairs are conditioned on the same class label and share 
the same starting noise realization $x_T$, enabling a direct visual comparison between 
the samplers.

\begin{table}[H]
   % \centering
    \caption{Theoretical number of function evaluations (NFE) per generated sample on ImageNet. We report forward passes (FPs) and backward passes (BPs) separately. Counts are computed per algorithm step, independent of accept/reject outcomes.}
    \label{tab:nfe_imagenet}
        \footnotesize
    \begin{tabularx}{\textwidth}{CCCC}
\toprule
         & \textbf{Sampler} & \textbf{NFE (FP)} & \textbf{NFE (BP)} \\
      \midrule
        \multirow{2}{*}{Score}
        & Reverse & 1000 & 0 \\
        & HMC-4C  & 19,981 & 0 \\
   \bottomrule
    \end{tabularx}
\end{table}
\vspace{-12pt}

\unskip

\subsection{Image Tapestry}
\label{sec:expdetails:image_tapestry}

A cosine schedule is used for the parameter $\beta_t$. The stepsize parameters in this case is simply $a=1$ and $b = 1$, i.e., $\langStep_t = \beta_t$.

\section{Additional Experiments}

\subsection{Sanity Check: Line vs. Curve Path on MNIST}
\label{sec:app:mnist_line_curve}

We conducted an additional experiment on the MNIST dataset to further assess 
the effect of path choice when estimating pseudo-energy differences. 
The setup followed the same spirit as in Appendix~\ref{sec:app:ped}. 
Specifically, we trained $10$ independent score models and sampled $2000$ states 
from the forward process at various timesteps $t$. 
From each state, we performed a single HMC proposal consisting of 
$3$ leapfrog steps with stepsize 
$\varepsilon_t = a \beta_t^b$, where $a=4.03$ and $b=1.34$ 
(as in the CIFAR-100 experiments). 
This produced a proposed state $\hat{x}_t$ for each sampled pair $(x_t, \hat{x}_t)$. 

We then computed the pseudo-energy difference between $x_t$ and $\hat{x}_t$ 
using two different paths:  
(i) a straight line between the points, with $n=10$ (Algorithm~\ref{alg:line}), and  
(ii) the curved path defined by the leapfrog steps themselves, with $m=3$ (Algorithm~\ref{alg:curve}). 

The discrepancy between the two estimates was measured using the symmetric \mbox{relative error}
\begin{equation}
\label{eq:path_discrepancy}
\frac{2 |\Delta E_{\text{line}} - \Delta E_{\text{curve}}|}
{|\Delta E_{\text{line}}| + |\Delta E_{\text{curve}}|}.
\end{equation}
For each model, we computed the median relative error across the $2000$ sampled pairs. 
We then reported the mean and standard deviation across the $10$ trained models, 
yielding a value of $0.022 \pm 0.002$. 
This small error indicates that line and curve integration paths produce 
very similar pseudo-energy differences, 
supporting the interpretation that the score behaves approximately conservatively 
on the local scales relevant for MCMC proposals.

{\subsection{Sanity Check: Line vs. Curve Path on CIFAR-100} \label{sec:appcifar100}}

{To assess path sensitivity in a higher-dimensional setting, we repeat the same analysis using the trained CIFAR-100 score model employed in our main experiments.}

{We sample $2000$ states from the forward diffusion process at various timesteps $t$. 
From each state, we perform a single HMC proposal consisting of $3$ leapfrog steps with step size 
$\varepsilon_t = a \beta_t^b$, using the parameters selected in Table \ref{tab:mcmc_stepsize_cifar100}. 
This produces pairs $(x_t, \hat{x}_t)$ analogous to the MNIST experiment.}

We compute the pseudo-energy difference between $x_t$ and $\hat{x}_t$ using both a straight-line path and the curved leapfrog path, and measure the symmetric relative discrepancy defined in Equation~\eqref{eq:path_discrepancy}.

Across the $2000$ sampled pairs, we obtain a median discrepancy of $0.038$. 
While slightly larger than the MNIST result ($0.022 \pm 0.002$), 
the discrepancy remains small relative to the magnitude of the pseudo-energy differences. 
This suggests that the score continues to behave approximately conservatively on the local scales relevant for our MCMC proposals, resulting in limited path sensitivity even in the higher-dimensional CIFAR-100 setting.

\vspace{12pt}
{\subsection{Integration Mesh Convergence on CIFAR-100}
 \label{sec:app_mesh}}
 
{To assess numerical stability of the line-integral approximation, we compare pseudo-energy differences computed with varying numbers of discretization points $n$ against a high-resolution reference with $n_{\text{ref}} = 40$.}

{We measure the symmetric relative discrepancy
\begin{equation}
\label{eq:mesh_discrepancy}
r_{\text{mesh}}(n) =
\frac{2 \left|\Delta E^{(n)}_{\text{line}} - \Delta E^{(40)}_{\text{line}}\right|}
{\left|\Delta E^{(n)}_{\text{line}}\right| + \left|\Delta E^{(40)}_{\text{line}}\right|}.
\end{equation}}

{Across $2000$ sampled pairs, we obtain median discrepancies
$r_{\text{mesh}}(5) = 0.032$,
$r_{\text{mesh}}(10) = 0.0064$, and
$r_{\text{mesh}}(20) = 8.2 \times 10^{-4}$.
These results indicate rapid convergence of the numerical integration scheme. 
In particular, for $n \ge 10$, the discretization error is already below $1\%$ relative to the high-resolution reference and becomes negligible for $n \ge 20$.}

\end{document}